\documentclass[sigconf]{acmart}
\AtBeginDocument{
  \providecommand\BibTeX{{
    \normalfont B\kern-0.5em{\scshape i\kern-0.25em b}\kern-0.8em\TeX}}}

\setcopyright{acmlicensed}
\copyrightyear{2024}
\acmYear{2024} 
\acmConference[MM '24]{Proceedings of the 32nd ACM International Conference on Multimedia}{October 28-November 1, 2024}{Melbourne, VIC, Australia}
\acmBooktitle{Proceedings of the 32nd ACM International Conference on Multimedia (MM '24), October 28-November 1, 2024, Melbourne, VIC, Australia}
\acmDOI{10.1145/3664647.3680667}
\acmISBN{979-8-4007-0686-8/24/10}

\usepackage{balance} 
 
\begin{document}
\title{MSTA3D: Multi-scale Twin-attention for 3D Instance Segmentation}

\author{Duc Dang Trung Tran}
\orcid{0009-0007-0061-4642}
\email{trandangtrungduc@seoultech.ac.kr}
\affiliation{
 \institution{Seoul National University of Science and Technology}
 \department{Department of Electrical and Information Engineering}
 \city{Seoul}
 \country{Republic of Korea}
}

\author{Byeongkeun Kang}
\orcid{0000-0003-2537-7720}
\email{byeongkeun.kang@seoultech.ac.kr}
\orcid{1234-5678-9012}
\affiliation{
 \institution{Seoul National University of Science and Technology}
 \department{Department of Electronic Engineering}
 \city{Seoul}
 \country{Republic of Korea}
}

\author{Yeejin Lee}
\authornote{Corresponding author}
\orcid{0000-0002-3439-5042}
\email{yeejinlee@seoultech.ac.kr}
\affiliation{
 \institution{Seoul National University of Science and Technology}
 \department{Department of Electrical and Information Engineering}
 \city{Seoul}
 \country{Republic of Korea}
}

\renewcommand{\shortauthors}{Duc Dang Trung Tran, Byeongkeun Kang and Yeejin Lee}

\newcommand{\etal}{\textit{et al}. }
\newcommand{\ie}{\textit{i}.\textit{e}., }
\newcommand{\eg}{\textit{e}.\textit{g}., }
\newcommand{\etc}{\textit{etc}.\:}

\providecommand{\eref}[1]{\eqref{#1}} 
\providecommand{\cref}[1]{Chapter~\ref{#1}}
\providecommand{\sref}[1]{Section~\ref{#1}}
\providecommand{\fref}[1]{Figure~\ref{#1}}
\providecommand{\tref}[1]{Table~\ref{#1}}
\providecommand{\thref}[1]{Theorem~\ref{#1}}

\providecommand{\R}{\ensuremath{\mathbb{R}}}
\providecommand{\C}{\ensuremath{\mathbb{C}}}
\providecommand{\E}{\ensuremath{\mathbb{E}}}
\providecommand{\N}{\ensuremath{\mathbb{N}}}
\providecommand{\Z}{\ensuremath{\mathbb{Z}}}

\providecommand{\bfP}{\mathbf{P}}

\providecommand{\abs}[1]{\lvert#1\rvert}
\providecommand{\norm}[1]{\lVert#1\rVert}
\providecommand{\inprod}[1]{\langle#1\rangle}
\providecommand{\bydef}{\overset{\text{def}}{=}}

\renewcommand{\vec}[1]{\ensuremath{\boldsymbol{#1}}}
\newcommand{\mat}[1]{\mathbf{#1}}

\providecommand{\mA}{\mat{A}}
\providecommand{\mB}{\mat{B}}
\providecommand{\mC}{\mat{C}}
\providecommand{\mD}{\mat{D}}
\providecommand{\mE}{\mat{E}}
\providecommand{\mF}{\mat{F}}
\providecommand{\mG}{\mat{G}}
\providecommand{\mH}{\mat{H}}
\providecommand{\mI}{\mat{I}}
\providecommand{\mJ}{\mat{J}}
\providecommand{\mK}{\mat{K}}
\providecommand{\mL}{\mat{L}}
\providecommand{\mM}{\mat{M}}
\providecommand{\mN}{\mat{N}}
\providecommand{\mO}{\mat{O}}
\providecommand{\mP}{\mat{P}}
\providecommand{\mQ}{\mat{Q}}
\providecommand{\mR}{\mat{R}}
\providecommand{\mS}{\mat{S}}
\providecommand{\mT}{\mat{T}}
\providecommand{\mU}{\mat{U}}
\providecommand{\mV}{\mat{V}}
\providecommand{\mW}{\mat{W}}
\providecommand{\mX}{\mat{X}}
\providecommand{\mY}{\mat{Y}}
\providecommand{\mZ}{\mat{Z}}

\begin{abstract}
    Recently, transformer-based techniques incorporating superpoints have become prevalent in 3D instance segmentation. However, they often encounter an over-segmentation problem, especially noticeable with large objects. Additionally, unreliable mask predictions stemming from superpoint mask prediction further compound this issue. To address these challenges, we propose a novel framework called MSTA3D. It leverages multi-scale feature representation and introduces a twin-attention mechanism to effectively capture them. Furthermore, MSTA3D integrates a box query with a box regularizer, offering a complementary spatial constraint alongside semantic queries. Experimental evaluations on ScanNetV2, ScanNet200 and S3DIS datasets demonstrate that our approach surpasses state-of-the-art 3D instance segmentation methods.
\end{abstract}



\maketitle

\section{Introduction}
Given 3D point clouds, 3D instance segmentation refers to a task that involves identifying and separating individual objects within a 3D scene, including detecting object boundaries and assigning a unique label to each identified object. Its significant role in computer vision has surged corresponding to the demand for 3D perception in various applications, such as augmented/virtual reality~\cite{park2020deep,lehtola2017comparison}, autonomous driving~\cite{yurtsever2020survey}, robotics~\cite{xie2021unseen}, and indoor scanning~\cite{li2022dn}.

\begin{figure}[t]
    \centering
    \includegraphics[width=0.9\linewidth]{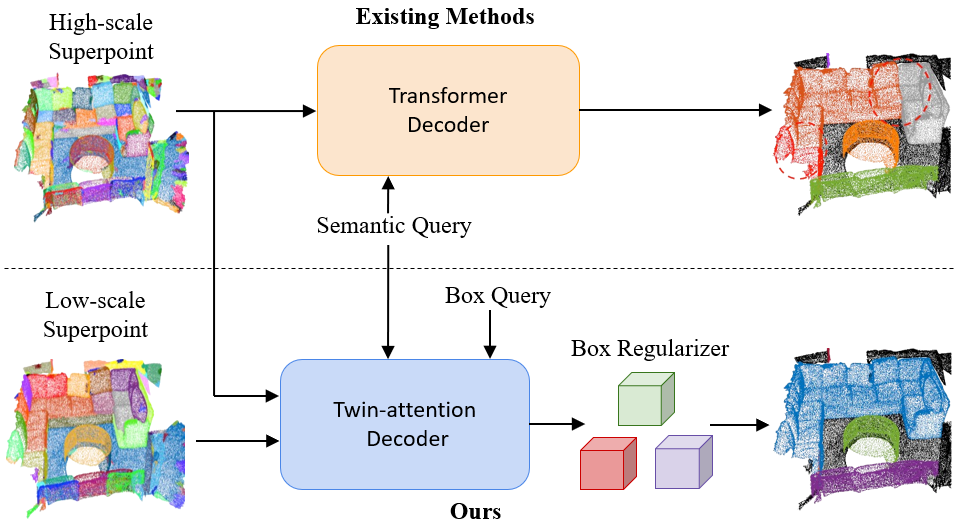}
    \caption{The proposed MSTA3D, a 3D instance segmentation framework, tackles existing challenges by leveraging multi-scale feature representation and spatial query/regularizer. \vspace{-0.3cm}} 
\end{figure}

In the literature, 3D instance segmentation methods are commonly categorized into four main approaches: proposal-based~\cite{sun2023neuralbf, yang2019learning, qi2017pointnet++, liu2020learning, engelmann20203d, kolodiazhnyi2024top, hou20193d}, grouping-based~\cite{lahoud20193d, jiang2020pointgroup, han2020occuseg, liang2021instance, chen2021hierarchical, dong2022learning, vu2022softgroup, zhong2022maskgroup, vu2023scalable, pham2019jsis3d}, kernel-based~\cite{he2021dyco3d, jiang2020pointgroup, he2022pointinst3d, wu20223d, ngo2023isbnet}, and transformer-based methods~\cite{sun2023superpoint, schult2023mask3d, lai2023mask, lu2023query}. Proposal-based methods begin by generating a 3D bounding box and then using it to segment into an instance mask. Grouping-based methods aggregate points into clusters using per-point features, such as semantic or geometric cues, and then segment instances based on these clusters. Kernel-based methods are similar to grouping-based techniques but treat each potential instance as a kernel for dynamic convolution. 
However, these methods require a high-quality proposal or clustering algorithm that heavily relies on per-point prediction, resulting in a significant demand for computational resources. 

To address these issues, transformer-based methods have been proposed, treating each potential instance as an instance query and refining it through a series of transformer decoder blocks. However, predicting instances from point clouds inherently presents substantial challenges due to their typically lacking clear structure, unlike the regular grid arrangement found in images. Moreover, managing large-scale input point clouds further requires costly computations and extensive memory resources. Thus, recent transformer approaches leverage superpoints, which roughly offer contextual relationships between object parts with reduced memory usage. Nevertheless, existing transformer-based approaches employing superpoints often suffer from performance degradation due to over-segmentation problems and the unreliability of mask predictions. These challenges are described as follows: 
(1) Existing methods are prone to over-segmentation, especially when dealing with large objects such as doors, curtains, bookshelves, and backgrounds. Additionally, converting labels from superpoints to points can introduce unreliability into the categorical grouping. (2) The learning process for point-wise classification and aggregation encounters challenges due to the sparse and irregular distribution of observed scene points. 

Hence, we propose a novel framework that leverages multi-scale superpoint features and simultaneously incorporates global/local spatial constraints. This framework is aimed at mitigating previously mentioned over-segmentation challenges and overcoming the limitations in point-wise classification. Specifically, to capture features at various scales, we generate superpoints at different scales, enabling effective feature representation of large objects and backgrounds as well as small objects. Correspondingly, we introduce a novel attention scheme, named twin-attention, to effectively fuse features at different scales. Moreover, we introduce the concept of box query, in addition to semantic query, which is trained using the proposed twin-attention decoder and refined by the spatial regularizer to enhance the confidence of mask predictions. 
Furthermore, we utilize this bounding box prediction to enhance the reliability of mask predictions during the inference phase.

In summary, the contributions of this study are as follows:
\begin{itemize}
\item We propose a twin-attention-based decoder for effectively representing multi-scale features to tackle over-segmentation challenges observed in large objects and backgrounds.
\item We introduce the notion of box query with box regularizer to provide complementary supervision without additional annotation requirements. This enforces spatial constraints for each instance during the query learning process, resulting in enhanced object localization and reduced background noise.
\item Extensive experiments are conducted on widely-used benchmark datasets, including ScanNetV2~\cite{dai2017scannet}, ScanNet200~\cite{rozenberszki2022language}, and S3DIS~\cite{armeni2016s3dis}, demonstrating the effectiveness of the proposed approach and achieving state-of-the-art results.
\end{itemize}

\section{Related Work}

Current approaches of instance segmentation on 3D point clouds can be classified into four categories: proposal-based, grouping-based, kernel-based, and transformer-based methods.

\noindent \textbf{Proposal-based methods.} In the early stages of this approach, 3D-BoNet~\cite{yang2019learning} was introduced, employing global characteristics derived from PointNet++~\cite{qi2017pointnet++} to generate bounding boxes. These bounding boxes were then integrated with point features to produce instance masks. GICN~\cite{liu2020learning} utilized a Gaussian distribution to estimate the center and size of each object instance for proposal prediction. Meanwhile, 3D-MPA~\cite{engelmann20203d} predicted instance centers and employed a graph convolutional network to group points around these centers, refining the features of proposals. 

\noindent \textbf{Grouping-based methods.} SSTNet~\cite{liang2021instance} represented the scene comprehensively by constructing a superpoint tree and traversing it to merge nodes, thus creating instance masks. PointGroup~\cite{jiang2020pointgroup} predicted the 3D displacement of each point from its instance’s center and identified clusters from both the original and center-shifted points. HAIS~\cite{chen2021hierarchical} introduced a hierarchical clustering technique, allowing smaller clusters to be either eliminated or merged into larger ones. SoftGroup~\cite{vu2022softgroup} permitted each point to be associated with multiple clusters representing diverse semantic classes to mitigate prediction inaccuracies. Softgroup++~\cite{vu2023scalable} extended SoftGroup to reduce computation time and search space for the clustering process.

\noindent \textbf{Kernel-based methods.} DyCo3D~\cite{he2021dyco3d} employed a clustering algorithm from PointGroup~\cite{jiang2020pointgroup} and lightweight 3D-UNet to generate kernels, while PointInst3D~\cite{he2022pointinst3d} opted for farthest-point sampling to produce kernels. DKNet~\cite{wu20223d} introduced candidate localization to yield more distinctive instance kernels. ISBNet~\cite{ngo2023isbnet} proposed a sampling-based instance-wise encoder to obtain a faster and more robust kernel for dynamic convolution. 

\noindent \textbf{Transformer-based methods.} SPFormer~\cite{sun2023superpoint} and Mask3D~\cite{schult2023mask3d} utilized the mask-attention to produce query instances based on voxel or superpoint features while MAFT~\cite{lai2023mask} improved efficiency by using position query instead of mask-attention. QueryFormer~\cite{lu2023query} optimized query instances with the query initialization module and proposes an affiliated transformer decoder to eliminate noise background.

\section{Box Guided Twin-attention Decoder for Multi-scale Superpoint}

In this section, we provide an in-depth discussion of the design choices underlying the proposed model. The problem addressed in this paper and the overall structure devised to tackle this problem are elucidated in \sref{sec:overview}. \sref{sec:backbone} explains the backbone network and the reasoning behind incorporating multi-scale superpoint features. \sref{sec:twin_attention_decoder} elaborates on the novel twin attention mechanism, and \sref{sec:box_regularizer} outlines the approach for constraining spatial regions for each instance. Finally, \sref{sec:training} provides a comprehensive explanation of each component of the loss function and the matching method employed.

\begin{figure*}[t]
    \centering
    \includegraphics[width=0.75\textwidth]{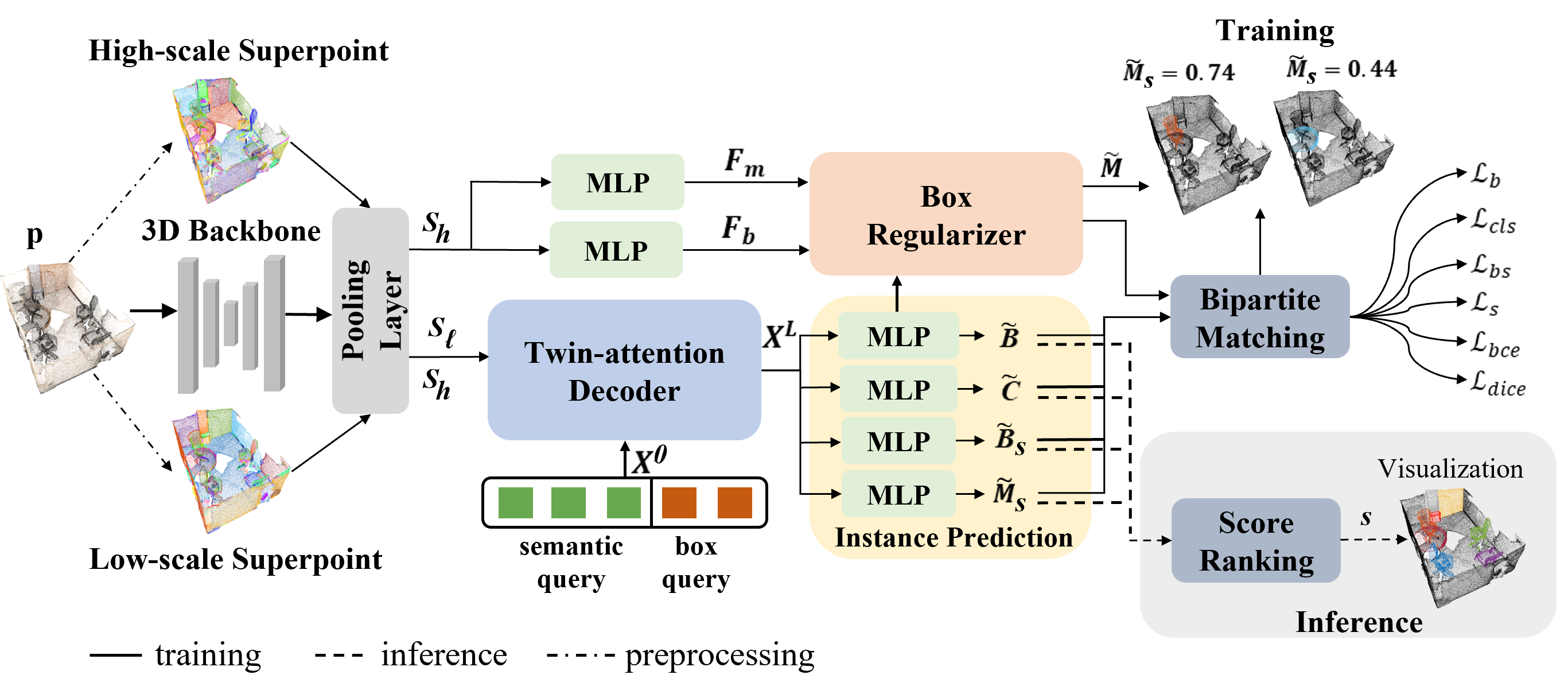}
    \vspace{-0.2cm}
    \caption{The MSTA3D framework for instance segmentation on point clouds. \vspace{-0.15cm}}
    \label{fig:overall_architecture}
\end{figure*}

\subsection{Architecture Overview} \label{sec:overview}
The goal of 3D instance segmentation is to predict precise point-wise boundaries of $N_I$ individual object instances given $N_h$ superpoints generated from point clouds, denoted as $\mathbf{p} \in\R^{N\times6}$. Each point cloud contains positional coordinates $(x, y, z)$ and color information $(r, g, b)$, where $N$ is the total number of points. A binary mask represents each of these $N_I$ instances, and a set of binary masks is collectively referred to $\mM \in\{0,1\}^{N_I\times N_h}$, 
where a value of $1$ indicates the presence of objects, and $0$ indicates their absence. Moreover, it is necessary to predict the semantic category associated with each instance, denoted as $\mC\in\Z^{N_I\times N_C}$, where $N_C$ is the total number of semantic categories.

To tackle the problem, we propose a model consisting of three key components, as illustrated in \fref{fig:overall_architecture}: (1) backbone network, (2) twin-attention decoder, and (3) box regularizer. First, the backbone network extracts multi-scale features $\mS_\ell $ and $\mS_h$ from the given inputs. These extracted features are then fed into the novel twin-attention decoder to generate instance proposals, represented as $\mX$,  guided by newly introduced box queries. Additionally, employing the box regularizer, the prediction of bounding boxes $\widetilde{\mB}$ along with their corresponding scores $\widetilde{\mB}_s$ are utilized to confine instance regions. Finally, through the twin-attention-based decoder and box regularizer, the final outputs include instance masks $\widetilde{\mathbf{M}}$ and corresponding labels $\widetilde{\mathbf{C}}$.

\subsection{Backbone Network} \label{sec:backbone}
As previously mentioned in \sref{sec:overview}, the backbone network, a 3D U-Net~\cite{ronneberger2015u, graham20183d}, takes voxelized point clouds as input and extracts point-wise features. The voxelization method utilized in this paper is based on the approach outlined in \cite{sun2023superpoint}.

We choose to utilize superpoint-wise labels instead of point-wise labels to reduce training time and memory consumption~(See \fref{fig:complexity_comparison}). Superpoints, as proposed by~\cite{landrieu2018large}, comprise multiple points sharing similar geometric properties and serve as a method to subsample point clouds. However, one potential drawback of using superpoints is the possibility of over-clustering scene~\cite{lu2023query, liang2021instance, sun2023superpoint, ngo2023isbnet}, which depends on the chosen grouping size. Therefore, we extract features from superpoints at multiple scales to ensure that the decoder accurately identifies objects of various sizes during the learning process. 

For this purpose, we pre-compute superpoint at two different scales by adjusting the number of neighbors for each point, as described in \cite{landrieu2018large}. Following this, a pooling layer is applied to aggregate information from the point-wise features into the multi-scale features. We then denote the outputs passing through this pooling layer as $\mS_\ell\in\R^{N_\ell\times D_b}$ for the lower-scale features and $\mS_h\in\R^{N_h\times D_b}$ for the higher-scale features. Here, $N_\ell$ and $N_h$ represent the number of low and high-scale superpoints, respectively, and $D_b$ is the embedding dimension from the backbone. 

\subsection{Twin-attention Decoder} \label{sec:twin_attention_decoder}

To leverage the advantages of multi-scale features $S_\ell$ and $S_h$ defined in \sref{sec:backbone}, we introduce a novel decoder structure consisting of a series of twin attentions. Inspired by~\cite{jiang2023dual,si2018dual,nam2017dual,fu2019dual}, the proposed twin-attention-based decoder is meticulously designed to integrate the multi-scale features. This decision is driven by the need to ensure that the proposed model effectively harnesses the diverse information presented across different scales of the input features. The twin decoder incorporates a spatial regularizer for guiding instance delineations. Further details regarding this spatial regularization are provided in the following section.

\noindent {\bf Region Constraint Instance Query.} We propose the concept of a box query, along with a semantic query~\cite{lai2023mask, lu2023query, sun2023superpoint}, to guide the regions of instance masks for more accurate predictions. This guidance enables the model to concentrate more effectively on regions of interest, potentially reducing instance region ambiguity and improving segmentation accuracy. Concretely, we construct a set of learnable instance queries $\mX^{0} \in \R^{N_o\times D_o}$ by concatenating the box queries with $6$ elements, denoted as $\mX_b \in \R^{N_o\times 6}$ and the semantic queries with $D_s$ elements, denoted as $\mX_s \in \R^{N_o\times D_s}$. The concatenation is expressed as $\mX^{0} = \left[ \mX_s; \mX_b\right]$, where $N_o$ denotes the number of queries~(proposals), initially set randomly with $N_o > N_I$, and  $D_o$ is the total dimensionality of a query vector~(\ie $D_o = D_s + 6$). Note that the number of instance proposals $N_o$ is configured to be larger than the ground truth $N_I$. Consequently, the proposals selected as the final ones are those with the highest confidence.

\begin{figure}[t]
    \centering
    \includegraphics[width=0.45\textwidth]{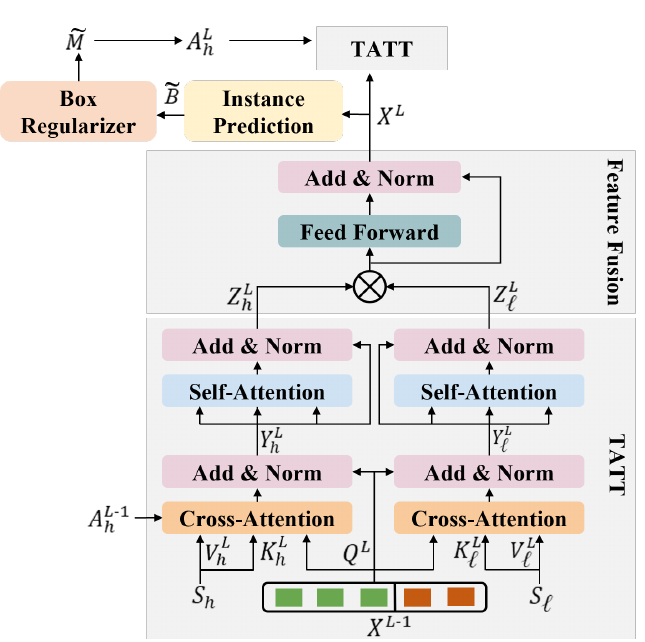}
    \caption{The architecture of twin-attention-based decoder. The twin-attention-based decoder fuses multi-scale features $\mS_h$ and $\mS_\ell$ and predicts $\mX^L$ by refining box queries. \vspace{-0.2cm}} \label{fig:twin_attention_decoder}
\end{figure}

\noindent {\bf Twin-Attention-Based Feature Extraction.} The proposed twin-attention-based decoder is structured to concurrently process low and high-scale features through weight-shared attention layers, as illustrated in \fref{fig:twin_attention_decoder}. It consists of a stack of six twin-attention blocks, indexed by $L$~($L = 1, \cdots, 6$), to process the outputs. Each block includes three sub-modules: cross and self-attention, feature fusion, and an instance prediction module. 

Let $\pi_{c}(\cdot)$ be the linear projection in the attention modules, where $c = \{q, v, k\}$ represents query, value, and key, respectively. In the cross-attention modules, the query matrix $\mQ^{L}\in \R^{N_o \times D_o}$ of the $L${\it -th} block is obtained after linearly projecting instance queries, computed as $\pi_q^{L}\left(\mX^{L-1}\right)$. Similarly, for low-scale feature $\mS_\ell$, the key matrix $\mK^{L}_\ell \in \R^{N_\ell \times D_o}$ and value matrix $\mV^{L}_\ell \in \R^{N_\ell \times D_o}$ are given by linearly projecting $\pi_k^{L}\left(\mS_\ell\right)$ and $\pi_v^{L}\left(\mS_\ell\right)$, respectively. Correspondingly, the key and value matrices of the high-scale features $\mS_h$ are derived as $\mK^{L}_h  = \pi_k^{L}\left(\mS_h\right)\in \R^{N_h \times D_o}$ and $\mV^{L}_h  = \pi_v^{L}\left(\mS_h\right) \in \R^{N_h \times D_o}$, respectively, using the same procedure. The proposed twin attention~(TATT) is then computed on the shared set of instance queries and multi-scale features to attend to relevant information, as follows:
\begin{subequations} 
\begin{align}
TATT(\mQ^{L}, \mK_h^{L}, \mV_h^{L})= & \sigma\left( \frac{Q^{L} {\left(K^{L}_h\right)}^{T}}{\sqrt{d}} + A^{L-1}_{h} \right) \cdot V_h^{L}, \label{eq:high_cross_attention}\\  
TATT(\mQ^{L}, \mK_\ell^{L}, \mV_\ell^{L}) = & 
\sigma\left( \frac{ Q^{L} \left(K_\ell^{L}\right)^T}{\sqrt{d}} \right)\cdot V_\ell^{L}, \label{eq:low_cross_attention}
\end{align}
\end{subequations}
where $\sigma(\cdot)$ denotes the row-wise softmax function, and $d$ is computed as the division of the dimension of the query~($D_o$) by the number of attention heads~(8 in the experiments of this study). For high-scale attention in \eref{eq:high_cross_attention}, we employ the superpoint mask attention $A^{L-1}_h \in \R^{N_o \times N_h}$ with a threshold $\tau = 0.5$, similar to the works in \cite{sun2023superpoint, lu2023query,schult2023mask3d}. However, we further enhance the mask attention prediction by incorporating the proposed box query and box regularizer, as elaborated in \sref{sec:box_regularizer}.

Next, denoting the outputs of the cross-attention modules passing through a residual connection~\cite{he2016deep} and layer norm~\cite{ba2016layer, vaswani2017attention} as $\mY_\ell^{L} \in \R^{N_o \times D_o}$ and $\mY_h^{L} \in \R^{N_o \times D_o}$, the subsequent self-attention layer computes twin attention on a set of linearly projected queries, keys, and values, in a similar manner to \eref{eq:high_cross_attention} and \eref{eq:low_cross_attention}, without applying mask attention. The computed outputs of the self-attention layer, denoted as $\mZ_\ell^{L} \in \R^{N_o \times D_o}$ and $\mZ_h^{L} \in \R^{N_o \times D_o}$, are then fed into the multi-scale feature fusion module to abstract both low-scale and high-scale contextual information. To accomplish this, we utilize element-wise multiplication to merge the features extracted from inputs of different scales, followed by further aggregation using a feedforward layer~(FFN), as follows:
\begin{equation}
\mX^{L} = FFN\left( \mZ_\ell^{L} \otimes \mZ_h^{L}\right), 
\end{equation}
where $\otimes$ denotes element-wise multiplication. The output $\mX^{L} \in \R^{N_o \times D_o}$ of this feature fusion module for $L$-th block is then utilized for the inputs of the next twin-attention blocks and box regularizer. In the training, the proposed twin-attention blocks are sequentially trained using an iterative prediction strategy~\cite{sun2023superpoint}. The output of the last decoder block~($\mX^6$) serves as the final instance proposals during inference. 

We have found that this introduced twin-attention-based decoder indeed enhances the performance of instance segmentation by leveraging multi-scale features. It enables the representation of objects of various sizes and captures the background details of the entire scene, effectively mitigating the possibility of over-segmentation~(See \tref{tab:loss_scale_study} and Appendix \ref{app:qualitative_results}).  

\noindent {\bf Instance and Box Prediction:} Given the fused feature output $\mX^{L}$ of the $L$-th block, the instance prediction module computes the mask score $\widetilde{\mM}_s \in \R^{N_o}$ for each instance and its corresponding class $\widetilde{\mC} \in \R^{N_o \times N_C}$ through multilayer perceptron~(MLP). These MLP layers employ linear activation for the mask score and softmax activation for classification. Moreover, to fully exploit the information from latent instance queries, we introduce two additional regressors with linear activation: $\widetilde{\mB} \in \R^{N_o \times 6}$ of instance box predictions and the corresponding box scores $\widetilde{\mB}_s \in \R^{N_o}$. 

These spatial feature-associated regressors are introduced without requiring additional annotation efforts. The proposed framework autonomously derives box information from the instance labels. Specifically, we choose axis-aligned bounding boxes due to the simplicity of constructing ground truths $\mB \in \R^{N_I \times 6}$ from existing instance annotations. Each bounding box is defined by six coordinates, representing the minimum and maximum 3D coordinates~(\ie $\left[ x_{min}, y_{min}, z_{min}, x_{max}, y_{max}, z_{max}\right]^{T}$) that enclose the instance. We observe that adopting this approach improves the overall performance of the proposed method because it utilizes bounding boxes as local spatial references for the box regularizer. These references provide contextual information about the local space occupied by each object relative to the entire scene during training, helping the model generate high-quality superpoint masks.

\subsection{Spatial Constraint Regularizer} \label{sec:box_regularizer}
\begin{figure}[t]
    \centering
    \includegraphics[width=0.95\linewidth]{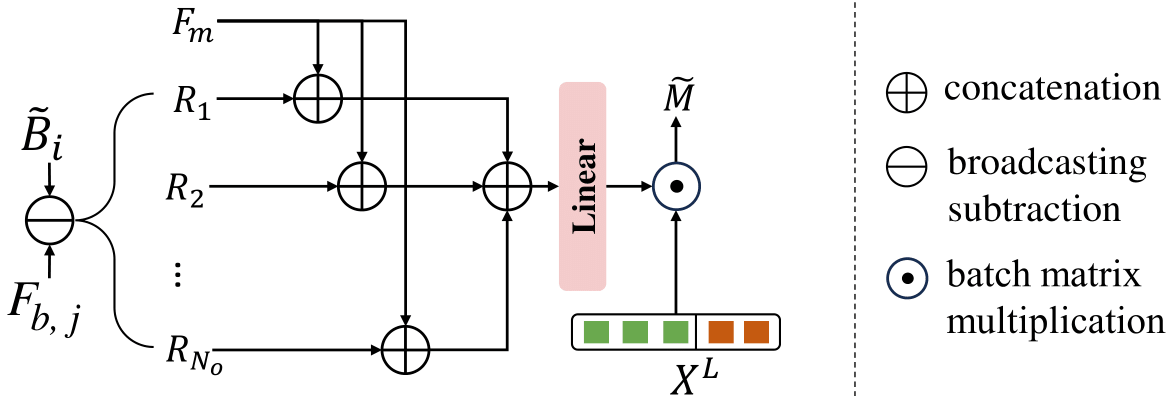}
    \caption{The architecture of box regularizer. The box regularizer predicts positional differences between bounding boxes derived from scene-wise features and those derived from instance-wise features.  \vspace{-0.3cm}} \label{fig:box_regularizer}
\end{figure}

In addition to incorporating box queries, we introduce the box regularizer to identify coarse object regions and constrain local spatiality. This regularization aims to enhance the representation of spatial latent features, utilizing box queries passed through the proposed twin-attention, as demonstrated in \tref{tab:loss_scale_study}.

The proposed box regularizer takes the box predictions $\widetilde{\mB}$ for each instance from the twin-attention-based decoder as input, along with features comprising scene-wise semantic score $\mF_m \in \mathbb{R}^{N_h \times D_s}$ for each superpoint and scene-wise box information $\mF_b \in \mathbb{R}^{N_h \times 6}$. These scene-wise features are obtained by passing $S_h$ through multilayer perceptron layers with linear activation. By utilizing these features, the goal is to provide the box regularizer with comprehensive scene-associated features, ensuring effective representation of local instance regions. This approach guarantees the enhanced confidence of superpoint mask predictions by analyzing instance features in the context of the entire scene.
Concretely, the box regularizer integrates instance-specific positional features $\widetilde{\textbf{b}}_{i} \in \mathbb{R}^{6}$, which denotes $i$-th row of $\widetilde{\mB}$, and scene-wise positional features $\mF_{b}$ for all instances.
By employing the broadcasting subtraction operator, we compute the relative positional difference $\mR_{i} \in \mathbb{R}^{N_h \times 6}$ for each instance box in relation to the entire scene, as follows:   
\begin{equation}
\mR_{i} = \{\widetilde{\mathbf{b}}_i\}^{N_h}\ominus \mF_{b}, 
\end{equation}
where $\ominus$ denotes a broadcasting subtraction operator, and $\{\widetilde{\mathbf{b}}_i\}^{N_h}$ represents the stretched version of $\widetilde{\mathbf{b}}_i$ by $N_h$.

To predict the binary mask in each twin-attention block, we concatenate these relative positional features $\mR_{i}$ with the scene-wise semantic features $\mF_m \in \mathbb{R}^{N_h \times D_s}$ to generate new $N_o$ features that simultaneously capture spatial and semantic information. The predicted binary masks are computed through batch matrix multiplication, as follows:
\begin{equation}\label{eq:binary_mask_prediction}
\widetilde{\mM} = Linear \left( \left[ \{{\mR_i}\}_{i=1}^{N_o}; \{\mF_m\}^{N_o}\right] \right) \odot \mX^{L}.
\end{equation}
In \eref{eq:binary_mask_prediction}, $\{{\mR_i}\}_{i=1}^{N_o}$ constructs to a tensor in $\mathbb{R}^{N_o \times \left(N_h \times 6\right)}$, and $\{\mF_m\}^{N_o}$ represents the stretched version of $\{\mF_m\}$ by $N_o$ using broadcasting operators, resulting in the concatenated features in $\mathbb{R}^{N_o \times \left(N_h \times D_o\right)}$ (where $D_o = D_s + 6$); $\odot$ denotes batch matrix multiplication, and $Linear(\cdot)$ denotes a linear activation function. 

Intuitively, the relative positional variation $\mR_{i}$ captures the prediction differences between the bounding box predicted from scene-wise global dependency and the estimated box derived from instance-wise local dependency. This suggests that mask prediction can be enhanced by abstracting both scene-wise features and individual instance-wise features. Moreover, this prediction of relative positional disparities can offer valuable feedback to the model, enabling it to refine its predictions based on discrepancies observed at different scales. 

\begin{table*}
  \setlength{\tabcolsep}{1.5pt}
  \caption{Performance comparisons of 3D instance segmentation on ScanNetV2 hidden test set in terms of the mean average precision and mAP$_{25}$ scores for each class \vspace{-0.3cm}}
  \label{tab:comparison_of_ScanNet}
  \begin{tabular}{@{}l@{}|@{}c@{}|ccc|cccccccccccccccccc}
    \toprule
    Methods & Venue & mAP &  mAP$_{50}$ & mAP$_{25}$ & \rotatebox[origin=c]{90}{bath} & \rotatebox[origin=c]{90}{bed} & \rotatebox[origin=c]{90}{bkshf} & \rotatebox[origin=c]{90}{cabinet} & \rotatebox[origin=c]{90}{chair} & \rotatebox[origin=c]{90}{counter} & \rotatebox[origin=c]{90}{curtain} & \rotatebox[origin=c]{90}{desk} & \rotatebox[origin=c]{90}{door} & \rotatebox[origin=c]{90}{other} & \rotatebox[origin=c]{90}{picture} & \rotatebox[origin=c]{90}{fridge} & \rotatebox[origin=c]{90}{s. cur.} & \rotatebox[origin=c]{90}{sink} & \rotatebox[origin=c]{90}{sofa} & \rotatebox[origin=c]{90}{table} & \rotatebox[origin=c]{90}{toilet} & \rotatebox[origin=c]{90}{window}\\
    \midrule
    DyCo3D~\cite{he2021dyco3d}        & CVPR 21    & 39.5          & 64.1          & 76.1 & 100 & 93.5 & 89.3 & 75.2 & 86.3 & 60.0 & 58.8 & 74.2 & 64.1 & 63.3 & 54.6 & 55.0 & 85.7 & 78.9 & 85.3 & 76.2 & 98.7 & 69.9 \\
    SSTNet~\cite{liang2021instance}   & ICCV 21    & 50.6       
    & 78.9            & 69.8 & 100 & 84.0 & 88.8 & 71.7 & 83.5 & 71.7 & 68.4 & 62.7 & 72.4 & 65.2 & 72.7 & 60.0 & 100 & 91.2 & 82.2 & 75.7 & 100 & 69.1 \\
    HAIS~\cite{chen2021hierarchical}  & ICCV 21    & 45.7          & 69.9          & 80.3 & 100 & 99.4 & 82.0 & 75.9 & 85.5 & 55.4 & 88.2 & 82.7 & 61.5 & 67.6 & 63.8 & 64.6 & 100 & 91.2 & 79.7 & 76.7 & 99.4 & 72.6 \\
    DKNet~\cite{wu20223d}             & ECCV 22    & 53.2          & 71.8          & 81.5 & 100 & 93.0 & 84.4 & 76.5 & 91.5 & 53.4 & 80.5 & 80.5 & 80.7 & 65.4 & 76.3 & 65.0 & 100 & 79.4 & 88.1 & 76.6 & 100 & 75.8 \\
    PBNet~\cite{zhao2023divide}       & ICCV 23    & 57.3          & 74.7        & 82.5 & 100 & 96.3 & 83.7 & 84.3 & 86.5 & 82.2 & 64.7 & 87.8 & 73.3 & 63.9 & 68.3 & 65.0 & 100 & 85.3 & 87.0 & 82.0 & 100 & 74.4 \\
    TD3D~\cite{kolodiazhnyi2024top}   & WACV 24    & 48.9          & 75.1          & 87.5 & 100 & 97.6 & 87.7 & 78.3 & 97.0 & 88.9 & 82.8 & 94.5 & 80.3 & 71.3 & 72.0 & 70.9 & 100 & 93.6 & 93.4 & 87.3 & 100 & 79.1 \\
    SoftGroup~\cite{vu2022softgroup}  & CVPR 22    & 50.4          & 76.1          & 86.5 & 100 & 96.9 & 86.0 & 86.0 & 91.3 & 55.8 & 89.9 & 91.1 & 76.0 & 82.8 & 73.6 & 80.2 & 98.1 & 91.9 & 87.5 & 87.7 & 100 & 82.0 \\ 
    ISBNet~\cite{ngo2023isbnet}       & CVPR 23    & 55.9          & 75.7          & 83.5 & 100 & 95.0 & 73.1 & 81.9 & 91.8 & 79.0 & 74.0 & 85.1 & 83.1 & 66.1 & 74.2 & 65.0 & 100 & 93.7 & 81.4 & 83.6 & 100 & 76.5 \\
    SPFormer~\cite{sun2023superpoint} & AAAI 23    & 54.9          & 77.0          & 85.1 & 100 & 99.4 & 80.6 & 77.4 & 94.2 & 63.7 & 84.9 & 85.9 & 88.9 & 72.0 & 73.0 & 66.5 & 100 & 91.1 & 86.8 & 87.3 & 100 & 79.6\\
    Mask3D~\cite{schult2023mask3d}    & ICRA 23    & 55.2          & 78.0          & 87.0 & 100 & 98.5 & 78.2 & 81.8 & 93.8 & 76.0 & 74.9 & 92.3 & 87.7 & 76.0 & 78.5 & 82.0 & 100 & 91.2 & 86.4 & 87.8 & 98.3 & 82.5  \\
    MAFT~\cite{lai2023mask}           & ICCV 23    & \textbf{59.6} & 78.6          & 86.0 & 100 & 99.0 & 81.0 & 82.9 & 94.9 & 80.9 & 68.8 & 83.6 & 90.4 & 75.1 & 79.6 & 74.1 & 100 & 86.4 & 84.8 & 83.7 & 100 & 82.8 \\
    QueryFormer~\cite{lu2023query}    & ICCV 23    & 58.3          & 78.7          & 87.4 & 100 & 97.8 & 80.9 & 87.6 & 93.6 & 70.2 & 71.6 & 92.0 & 87.5 & 76.6 & 77.2 & 81.8 & 100 & 99.5 & 91.6 & 89.2 & 100 & 76.7 \\
    
    \textbf{MSTA3D (Ours)}             &     -    & 56.9          & \textbf{79.5} & \textbf{87.9} & 100 & 99.4 & 92.1 & 80.7 & 93.9 & 77.1 & 88.7 & 92.3 & 86.2 & 72.2 & 76.8 & 75.6 & 100 & 91.0 & 90.4 & 83.6 & 99.9 & 82.4\\ 
  \bottomrule
\end{tabular}
\end{table*}

\subsection{Training and Inference} 
\label{sec:training}
\noindent {\bf Training: } To train the proposed framework, we formulate the loss function $\mathcal{L}$, as follows:
\begin{equation}
\mathcal{L} =  
 \beta_{m} \cdot(\mathcal{L}_{bce} +\mathcal{L}_{dice}) + \beta_{cls} \cdot\mathcal{L}_{cls} +
 \beta_{s}\cdot\mathcal{L}_s + \beta_{b}\cdot\mathcal{L}_{b} + \beta_{bs}\cdot\mathcal{L}_{bs}
\end{equation} 
where the coefficients $\beta_{m}$, $\beta_{cls}$, $\beta_{s}$, $\beta_{b}$, and $\beta_{bs}$ determine the contribution of each loss function to the total loss function. The classification loss $\mathcal{L}_{cls}$ is defined by the multi-class cross-entropy loss for object class categorization, and mask prediction utilizes a combination of binary cross-entropy $\mathcal{L}_{bce}$ and dice loss $\mathcal{L}_{dice}$. Additionally, we use L2 loss $\mathcal{L}_{s}$ for predicting mask scores used for matching during training, L1 loss $\mathcal{L}_{b}$ for box coordinate prediction, and L2 loss $\mathcal{L}_{bs}$ for box score prediction, as follows:
\begin{subequations}
\begin{align}
\mathcal{L}_s = & \frac{1}{ \lVert \varmathbb{1}_{iou_{\scriptscriptstyle ms}}\rVert_{1} } \cdot \bigg\lVert \varmathbb{1}_{iou_{\scriptscriptstyle ms}} \otimes \left( \widetilde{\mM}_{s} - iou_{\scriptscriptstyle ms}\right) \bigg\rVert_{2}^{2}, \vspace{0.3cm}\\
\mathcal{L}_{b} = & \frac{1}{N_I} \big\lVert \widetilde{\mB} - \mB \big\rVert_{1,1}, \vspace{0.3cm}\\
\mathcal{L}_{bs} = & \frac{1}{ \lVert \varmathbb{1}_{iou_{\scriptscriptstyle bs}}\rVert_{1} } \cdot \bigg\lVert \varmathbb{1}_{iou_{\scriptscriptstyle bs}} \otimes \left( \widetilde{\mB}_{s} - iou_{\scriptscriptstyle bs}\right) \bigg\rVert_{2}^{2},
 \end{align}
\end{subequations}
where $||\cdot||_{p}$ denotes the vector $p$-norm, and $||\cdot||_{p,q}$ denotes the entry-wise matrix $p,q$-norm; $\varmathbb{1}_{\{iou_{\scriptscriptstyle ms}\}}$ indicates whether the Intersection over Union~(IoU) between mask prediction and assigned ground truth is higher than a threshold $\eta_{ms}$; and $\varmathbb{1}_{\{iou_{\scriptscriptstyle bs}\}}$ indicates whether IoU between box prediction and assigned ground truth is higher than a threshold $\eta_{bs}$. 
In the experiments, we used $\beta_{cls}=0.5$ for class classification, $\beta_{s}=0.5$ for mask score prediction, $\beta_{m}=\beta_{b} = 1$ for mask and box prediction, $\beta_{bs} = 0.5$ for box score prediction, and $\eta_{ms}=\eta_{bs}=0.5$.

To assign proposals to each ground truth instance, we formulate a pairwise matching cost $C_{ij}$ to assess the similarity between the $i$-th proposal and the $j$-th ground truth. The matching cost $C_{ij}$ is defined as~\cite{sun2023superpoint}:
\begin{equation}
C_{ij} = -\lambda_{cls} . p_{i,c_j} + \lambda_{mask}C^{mask}_{ij},
\end{equation}
where $p_{i,c_j}$ denotes the $c_j$-th semantic category probability of the $i$-th proposal; $C^{mask}_{ij}$ denotes superpoint mask matching score; $\lambda_{cls}$ and $\lambda_{mask}$ are the coefficients of each term respectively. In the experiments, we set $\lambda_{cls} = 0.5$ and $\lambda_{mask} = 1$. 
The superpoint mask matching cost $C^{mask}_{ij}$ is calculated based on a binary cross-entropy~(BCE) and a dice loss with a Laplace smoothing~\cite{milletari2016v}:

\begin{equation}
C^{mask}_{ij} = BCE\left( \widetilde{\mathbf{m}}_{i}, \mathbf{m}_{j} \right) + \frac{\widetilde{\mathbf{m}}_{i}\cdot\mathbf{m}_{j}^{T}+1}{\lVert \widetilde{\mathbf{m}}_{i}\rVert_{1} + \lVert \mathbf{m}_{j} \rVert_{1}+1},
\end{equation} 
where $\widetilde{\mathbf{m}_{i}}$ is the $i$-th row of the superpoint mask prediction $\widetilde{\mM}$, and $\mathbf{m}_{j}$ is the $j$-th row of the ground truth mask $\mM$. 
Additionally, we treat the task of matching proposals with ground truth as an optimal assignment problem. Hence, we employ the Hungarian algorithm~\cite{kuhn1955hungarian} to identify the optimal solution for this task. Once completing the assignment, we categorize the proposals not assigned to ground truth as the ``no instance'' class. 

\noindent \textbf{Inference.} During inference, we compute instance confidence, considering that the number of proposals $N_o$ can be greater than the number of ground truth instances $N_I$. The confidence score of $i$-th proposal is computed based on classification probability $\widetilde{\mathbf{c}}_{i}$, IoU box score $\widetilde{b}_{s,i}$, IoU score $\widetilde{m}_{s,i}$ of corresponding superpoint mask.
Follow~\cite{sun2023superpoint,lai2023mask}, we utilize a superpoint mask score $a_i$ by computing by averaging the probabilities of superpoint that have a probability greater than 0.5 in each superpoint mask prediction. The final confidence score for each instance is defined as $\{s\}^{N_o}_{i=1}= \left\{{\widetilde{\mathbf{c}}_{i}} \cdot \widetilde{b}_{s,i} \cdot \widetilde{m}_{s,i} \cdot a_i \right\}_{i=0}^{N_o}$
and utilized for ranking predicted instances. In this work, we do not employ post-processing~(non-maximum suppression or DBSCAN) as~\cite{lu2023query, schult2023mask3d,ngo2023isbnet} to enhance inference speed.  

\section{Experimental Results}
In this section, we introduce the datasets and evaluation metrics utilized to validate the effectiveness of the proposed model in \sref{sec:dataset}, and implementation details are described in \sref{sec:implementation}. Following this, performance comparisons are presented in \sref{sec:results}, and analyses of additional studies on the model are demonstrated in \sref{sec:ablation_study}.

\subsection{Dataset and Evaluation Metric}
\label{sec:dataset}
\textbf{Dataset: } To validate the effectiveness of the proposed framework, we conducted experiments on the ScanNetV2 dataset \cite{dai2017scannet}, ScanNet200 dataset \cite{rozenberszki2022language}, and S3DIS dataset \cite{armeni2016s3dis}.

\begin{table}
  \caption{Performance comparisons of 3D instance segmentation on ScanNetV2 validation set in terms of mean average precision \vspace{-0.3cm}}
  \label{tab:comparison_of_scannet_val}
  \renewcommand{\tabcolsep}{4.0mm}{ {\renewcommand{\arraystretch}{1.0}
  \begin{tabular}{l|ccc}
    \toprule
    Methods                           & mAP           &  mAP$_{50}$     & mAP$_{25}$ \\
    \midrule
    DyCo3D~\cite{he2021dyco3d}        & 35.4          & 57.6          & 72.9 \\
    HAIS~\cite{chen2021hierarchical}  & 43.5          & 64.4          & 75.6 \\
    DKNet~\cite{wu20223d}             & 50.8          & 66.7          & 76.9 \\
    SoftGroup~\cite{vu2022softgroup}  & 46.0          & 67.6          & 78.9 \\
    PBNet~\cite{zhao2023divide}       & 54.3          & 70.5          & 78.9 \\
    TD3D~\cite{kolodiazhnyi2024top}   & 47.3          & 71.2          & 81.9 \\
    ISBNet~\cite{ngo2023isbnet}      & 54.5          & 73.1          & 82.5 \\
    Mask3D~\cite{schult2023mask3d}    & 55.2          & 73.7          & 83.5 \\
    SPFormer~\cite{sun2023superpoint} & 56.3          & 73.9          & 82.9 \\
    QueryFormer~\cite{lu2023query}    & 56.5          & 74.2          & 83.3 \\
    MAFT~\cite{lai2023mask}           & 58.4          & 75.6          & 84.5  \\ 
    \textbf{MSTA3D (Ours)}   & \textbf{58.4} & \textbf{77.0} & \textbf{85.4} \\
  \bottomrule
\end{tabular} \vspace{-0.3cm} }}
\end{table} 

ScanNetV2 comprises $1613$ scans, with $1201$ scans for training, $312$ for validation, and $100$ for testing. It includes $18$ object categories commonly used for 3D instance segmentation evaluation and $2$ background classes.

ScanNet200~\cite{rozenberszki2022language} extends ScanNetV2~\cite{dai2017scannet} with fine-grained categories and includes $198$ object categories used for 3D instance segmentation with an additional $2$ background classes. The dataset division into training, validation, and testing sets mirrors the original ScanNetV2 dataset.

S3DIS is a large-scale dataset that includes consists of $271$ scenes across 6 different areas, with each area containing $13$ categories.  For the experiments in this paper, evaluations were conducted using datasets of Area $5$.

\noindent \textbf{Evaluation Metric: } The evaluation of 3D instance segmentation typically relies on task-mean average precision~(mAP), a widely-used metric that averages scores across various IoU thresholds ranging from $50\%$ to $95\%$, incremented by $5\%$. 
mAP$_{50}$ and mAP$_{25}$ represent the scores at IoU thresholds of $50\%$ and $25\%$, respectively. Our evaluation of the ScanNetV2 and ScanNet200 datasets included mAP, mAP$_{50}$, and mAP$_{25}$ metrics, and we also utilized mean precision~(mPrec) and mean recall~(mRec) in our analysis of the S3DIS dataset.

\subsection{Implementation Details} \label{sec:implementation}
We implemented the proposed model using the PyTorch deep learning framework~\cite{paszke2017automatic} and conducted training with the AdamW optimizer~\cite{loshchilov2017decoupled}. The training was performed on a single A100 GPU with $40$GB of memory. We used an initial learning rate of $0.0001$, weight decay of $0.05$, and batch size of $2$, and employed a polynomial scheduler with a base of $0.9$ for $512$ epochs. 
For data augmentation, we applied random scaling, elastic distortion, random rotation, horizontal flipping around the z-axis, and random scaling for $x$, $y$, and $z$ coordinates. Additionally, we utilized noise and brightness augmentation for red, green, and blue colors.
Both ScanNetV2 and ScanNet200 datasets were processed with a voxel size of $2cm$. For S3DIS dataset, we used a voxel size of $5cm$. The proposed model adopted the same backbone design described in \cite{sun2023superpoint, lai2023mask}, resulting in a feature map with $32$ channels~($D_b = 32$). During training, scenes were limited to a maximum of $250,000$ points. During inference, the entire scenes were inputted into the network without any cropping, and the top 100 instances were selected based on their highest scores.

\begin{table}
  \setlength{\tabcolsep}{4.5pt}
  \caption{Performance comparisons of 3D on ScanNet200 validation set. The asterisk(*) indicates reproduced results. \vspace{-0.2cm}}
  \label{tab:comparison_of_scannet200}
  \renewcommand{\tabcolsep}{2.2mm}{ {\renewcommand{\arraystretch}{1.0}
  \begin{tabular}{l|ccc}
    \toprule
    Methods & mAP &  mAP$_{50}$ & mAP$_{25}$ \\
    \midrule 
    PointGroup~\cite{jiang2020pointgroup}                & -            & 24.5        & - \\
    PointGroup + LGround~\cite{rozenberszki2022language} & -            & 26.1        & - \\
    ISBNet ~\cite{ngo2023isbnet}                         & 24.5          & 32.7       &  -   \\
    SPFormer*~\cite{sun2023superpoint}                    & 25.2          & 33.8       & 39.6 \\
    TD3D* ~\cite{kolodiazhnyi2024top}                     & 23.1	         & 34.8	      & \textbf{40.4} \\
    \textbf{MSTA3D (Ours)}                                & \textbf{26.2} & \textbf{35.2}      & 40.1 \\
  \bottomrule
\end{tabular} \vspace{-0.2cm} } }
\end{table}

\begin{table}
  \caption{Performance comparisons of 3D instance segmentation on Area 5 of the S3DIS dataset \vspace{-0.3cm}}
  \label{tab:comparison_of_s3dis}
  \renewcommand{\tabcolsep}{4.0mm}{ {\renewcommand{\arraystretch}{1.0}
  \begin{tabular}{l|ccc}
    \toprule
    Methods  & mAP$_{50}$ &  mPrec & mRec \\
    \midrule
    DyCo3D~\cite{he2021dyco3d} & - & 64.3 & 64.2 \\
    DKNet~\cite{wu20223d} & - & 70.8 &  65.3  \\
    HAIS~\cite{chen2021hierarchical} & - & 71.1 &  65.0\\
    SoftGroup~\cite{vu2022softgroup} & 66.1 & 73.6 & 66.6 \\
    TD3D~\cite{kolodiazhnyi2024top} & 65.1 & 74.4 & 64.8 \\
    PBNet~\cite{zhao2023divide} & 66.4 & 74.9 & 65.4 \\
    SPFormer~\cite{sun2023superpoint} & 66.8 & 72.8 & 67.1 \\
    MAFT~\cite{lai2023mask} & 69.1 & - & - \\ 
    QueryFormer~\cite{lu2023query} & 69.9 & 70.5 & \textbf{72.2}  \\
    Mask3D~\cite{schult2023mask3d} & \textbf{71.9} & 74.3 & 63.7 \\
    \textbf{MSTA3D (Ours)} & 70.0 & \textbf{80.6} &  70.1 \\
  \bottomrule
\end{tabular} \vspace{-0.4cm}}}
\end{table} 

\subsection{Experimental Results} \label{sec:results}
\noindent \textbf{ScanNetV2.} The instance segmentation results of the ScanNetV2 test and validation sets are presented in Tables \ref{tab:comparison_of_ScanNet}\footnote{Note that the results under the name "TST3D" presented were obtained from the ScanNet Benchmark.} 
and \ref{tab:comparison_of_scannet_val}. Overall, the proposed model shows a significant improvement in mAP compared to previous studies, indicating its capability to capture intricate details and generate high-quality instance segmentation. In particular, compared to SPFormer \cite{sun2023superpoint}, the proposed model achieved a performance increase of +2.0 mAP, +2.5 mAP$_{50}$, and +2.8 mAP$_{25}$. Compared to Mask3D \cite{schult2023mask3d}, the improvement was +1.7 mAP, +1.5 mAP$_{50}$, and +0.9 mAP$_{25}$. Compared to MAFT \cite{lai2023mask}, the gains were +0.9 mAP$_{50}$ and +1.9 mAP$_{25}$. Finally, compared to QueryFormer \cite{lu2023query}, the increases were +0.8 mAP$_{50}$ and +0.5 mAP$_{25}$ on the hidden test set.
On the validation set, the proposed model outperformed other methods across all three metrics: mAP, mAP$_{25}$, and mAP$_{50}$. Specifically, compared to QueryFormer \cite{lu2023query}, the proposed model achieved a +1.9 mAP improvement, +2.8 mAP$_{50}$ improvement, and +2.1 mAP$_{25}$ improvement. Compared to MAFT \cite{lai2023mask}, the gains were +1.4 mAP$_{50}$ and +0.9 mAP$_{25}$. 
As evidenced by the mAP scores for each class, the proposed model exhibited superior performance, particularly for relatively large objects such as beds or bookshelves. Notably, it established a considerable lead of over 10\%  mAP$_{25}$ specifically for the bookshelf category. This demonstrates that our proposed method effectively captures large objects to mitigate over-segmentation. The qualitative performance comparisons are visualized in Appendix \ref{app:qualitative_results}.

\noindent \textbf{ScanNet200.} We also evaluated the proposed model on the ScanNet200 dataset using the validation set. As shown in \tref{tab:comparison_of_scannet200}, the results of the proposed model indicate a significant improvement compared to other methods. Specifically, the proposed model achieved an increase of +1 in mAP, +1.4 in mAP$_{50}$ and +0.5 in mAP$_{25}$ compared to SPFormer~\cite{sun2023superpoint}. Especially, compared to TD3D~\cite{kolodiazhnyi2024top}, our model achieved a +3.1 mAP and +0.4 mAP$_{50}$ improvement. Nevertheless, this dataset presents challenges for superpoint-wise labels due to its limited number of classes, indicating that point-wise methods will likely remain dominant. The qualitative performance comparisons are visualized in Appendix \ref{app:qualitative_results}.

\begin{table}
  \caption{Performance comparisons with varying the number of queries on the ScanNetV2 validation set \vspace{-0.2cm}}
  \label{tab:study_query_numbers}
  \renewcommand{\tabcolsep}{2.5mm}{ 
  {\renewcommand{\arraystretch}{1.0}
  \begin{tabular}{l|c|ccc}
    \toprule
    Methods & \# Queries & mAP           &  mAP$_{50}$     & mAP$_{25}$ \\
    \midrule 
    SPFormer~\cite{sun2023superpoint} & 100 & 54.2      & 72.4          & 82.8 \\
    SPFormer~\cite{sun2023superpoint} & 200 & 55.2      & 73.3          & 82.4 \\
    SPFormer~\cite{sun2023superpoint} & 400 & \underline{56.3}      & \underline{73.9}          & \underline{82.9} \\ 
    SPFormer~\cite{sun2023superpoint} & 800 & 55.9      & 73.7          & 83.8 \\ \hline
    MAFT~\cite{lai2023mask}     & 400 & 58.4      & 75.6          & 84.5 \\ \hline
    MSTA3D (Ours) & 150 & \underline{57.2}          & \underline{76.1}          & \underline{84.3} \\
    MSTA3D (Ours) & 200 & 57.4          & \underline{75.8}          & \underline{85.2} \\
    MSTA3D (Ours) & 250 & 58.2          & 76.9          & 85.4 \\
    \textbf{MSTA3D (Ours)} & 300 & 58.4          & \textbf{77.0} & \textbf{85.4} \\
    MSTA3D (Ours) & 350 & 58.6          & 76.6          & 85.2 \\
    MSTA3D (Ours) & 400 & 58.1          & 76.4          & 84.7 \\
    MSTA3D (Ours) & 450 & \textbf{58.9} & 76.7          & 85.0 \\
  \bottomrule
\end{tabular} \vspace{-0.1cm}}}
\end{table}

\begin{figure}[t]
    \centering
    \includegraphics[width=0.4\textwidth]{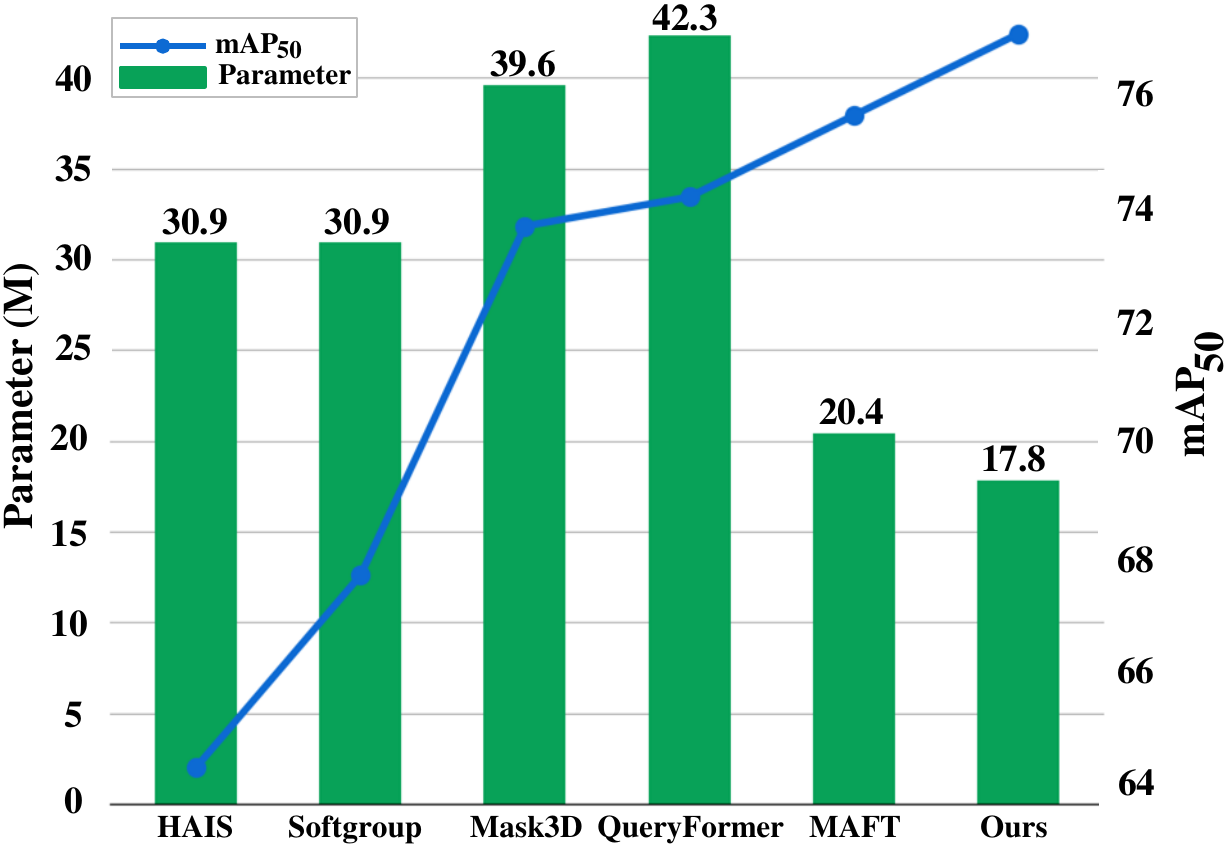}
    \caption{Comparisons of model complexity \vspace{-0.3cm}} \label{fig:complexity_comparison}
\end{figure}

\noindent \textbf{S3DIS.}  We conducted an evaluation of the proposed model on S3DIS using Area 5. As shown in \tref{tab:comparison_of_s3dis}, the results demonstrate slight improvements compared to previous works such as MAFT or QueryFormer. In comparison to SPFormer, we achieved a notable improvement of +3.2 mAP$_{50}$. Additionally, the proposed method outperformed other methods in terms of the mPrec metric.

\begin{table*}
    \centering
    \caption{Ablation study on different loss functions using both single-scale feature and the proposed multi-scale feature representation \vspace{-0.2cm}} \label{tab:loss_scale_study}
    {\renewcommand{\arraystretch}{1.1}
    \begin{tabular}{ccc|ccc|ccc|ccc}
    \toprule
    Instance & Box & Box & \multicolumn{3}{c|}{Low-scale only} & \multicolumn{3}{c|}{High-scale only} &  \multicolumn{3}{c}{Multi-scale} \\ \cline{4-12}
    Box & Score & Regularizer &  mAP & mAP$_{50}$ & mAP$_{25}$ & mAP & mAP$_{50}$ & mAP$_{25}$ & mAP & mAP$_{50}$ & mAP$_{25}$ \\ \hline
    -- & -- & -- & 47.2 & 67.4 & 78.0 & 56.3 & 73.9 & 82.9 & 56.4 & 75.0 & 84.0 \\  
    \checkmark & -- & -- & 47.3 & 67.1 & 78.8 & 56.7 & 74.6 & \textbf{84.6} & 57.0 & 74.8 & 84.4 \\    
    -- & -- & \checkmark & 48.0 & 68.0 & 79.6 & 56.2 & 74.6 & 83.5 & 56.9 & 75.5 & 84.3 \\   
    \checkmark & \checkmark & -- & 48.3 & 68.6 & 79.4 & 56.8 & 75.6 & 83.8 & 57.1 & 75.2 & 84.3 \\  
    \checkmark & -- & \checkmark & 48.1 & \textbf{68.9} & 79.9 & 57.4 & 75.0 & \textbf{84.6} & 57.5 & 75.4 & 84.8 \\  
    \checkmark & \checkmark & \checkmark & \textbf{49.4} & 68.0 & \textbf{81.2} & \textbf{57.6} & \textbf{75.6} & 84.5 & \textbf{58.4} & \textbf{77.0} & \textbf{85.4} \\   \bottomrule
    \end{tabular} }   
    \vspace{0.1cm} 
\end{table*}

\subsection{Ablation Study} \label{sec:ablation_study}
In this section, we conduct a thorough analysis to validate each component of the proposed approach, as outlined below. 

\noindent \textbf{Multi-scale Feature Representation and Loss Functions.} We assessed performance to further analyze the effectiveness of multi-scale feature representations and the proposed loss functions by disabling one or more components. The results of this study are tabulated in \tref{tab:loss_scale_study}. The results when utilizing instance bounding boxes confirmed that the introduced box queries improved the accuracy of instance recognition. Furthermore, utilizing bounding box features with box scores enhanced recognition accuracy even further. The introduced box regularizer also improved the accuracy of instance recognition, indicating the effectiveness of augmenting instance-wise spatial features with scene-wise spatial features from the superpoint-wise predictor.

When comparing the performance of multi-scale feature representation against single-scale feature representation, the results demonstrate that multi-scale superpoints enhanced the feature representation at different scales. This finding is consistent with the results presented in Table \ref{tab:comparison_of_ScanNet}, where the proposed model outperformed relatively large objects while performing well in recognizing small objects.

\noindent \textbf{Number of Queries.} We conducted a performance comparison of the proposed model using varying numbers of queries as shown in Table \ref{tab:study_query_numbers}. To ensure fairness in comparison, we exclusively evaluated models that utilize superpoint-wise information for both training and inference, namely SPFormer~\cite{sun2023superpoint} and MAFT~\cite{lai2023mask}. Remarkably, the proposed model achieved superior performance across all three metrics with only 150 queries (underlined score), compared to SPFormer's 400 queries (also underlined). Additionally, the proposed model outperformed MAFT in two metrics, mAP and mAP$_{50}$, with only 200 queries (underlined score), and achieved better results in three metrics with 250 queries. Note that the best scores among all methods are in bold.

\noindent \textbf{Model Complexity.} To validate the efficiency of the proposed model alongside the analysis of the number of queries, we compared the number of parameters of various methods. The results demonstrate that the proposed model achieved superior performance with significantly fewer parameters, specifically 24.5 million fewer than QueryFormer~\cite{lu2023query} and 2.6 million fewer than MAFT~\cite{lai2023mask}. Compared to the recent transformer-based method Mask3D~\cite{schult2023mask3d}, the proposed model utilized fewer than 21.8 million parameters, as evidenced in \fref{fig:complexity_comparison}. We also specifically compared the number of parameters between our work and MAFT~\cite{lai2023mask} in Appendix \ref{app:model_complexity}.

\section{Conclusion}
In this paper, we presented MSTA3D, a transformer-based method designed for 3D point cloud instance segmentation. To address the challenge of over-segmentation, particularly with background or large objects in the scene, we devised a multi-scale superpoint strategy. Furthermore, we introduced a twin-attention decoder to leverage both high-scale and low-scale superpoints simultaneously. This enhancement expands the model's capacity to capture features at various scales, thereby enabling better performance on large objects and reducing over-segmentation. In addition to the semantic query, we introduced the notion of a box query. This provides spatial context for generating high-quality instance proposals, assists the box regularizer in producing reliable instance masks, and contributes to box score regression, leading to significant performance improvements. Finally, we rigorously evaluated each of these components through extensive experiments.

\bibliographystyle{ACM-Reference-Format}
\balance


\newpage
\appendix

\counterwithin{table}{section}
\counterwithin{figure}{section}

\section*{Appendix}
\section{Qualitative Results} \label{app:qualitative_results}

In this supplementary material, we provide further qualitative results to validate the effectiveness of the proposed method compared to existing methods including ISBNet~\cite{ngo2023isbnet}, SPFormer~\cite{sun2023superpoint}, and MAFT~\cite{lai2023mask}. 

\begin{figure*}[!h]
    \begin{tabular}{@{}c@{}c@{}c@{}c@{}}
    \\
    Input & Low-scale Superpoint & High-scale Superpoint & Ground-truth \\
    \includegraphics[width=0.23\textwidth]{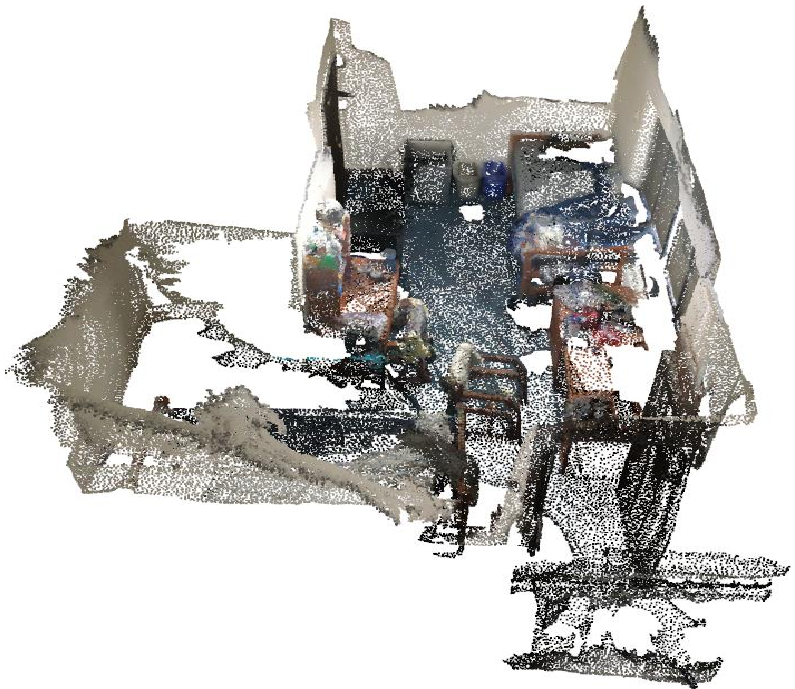} & \includegraphics[width=0.23\textwidth]{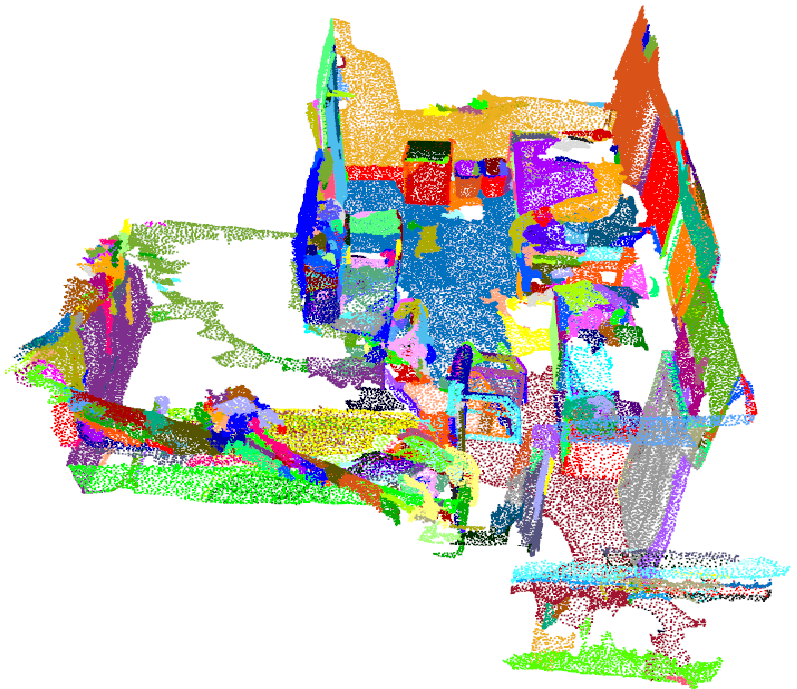} &  \includegraphics[width=0.23\textwidth]{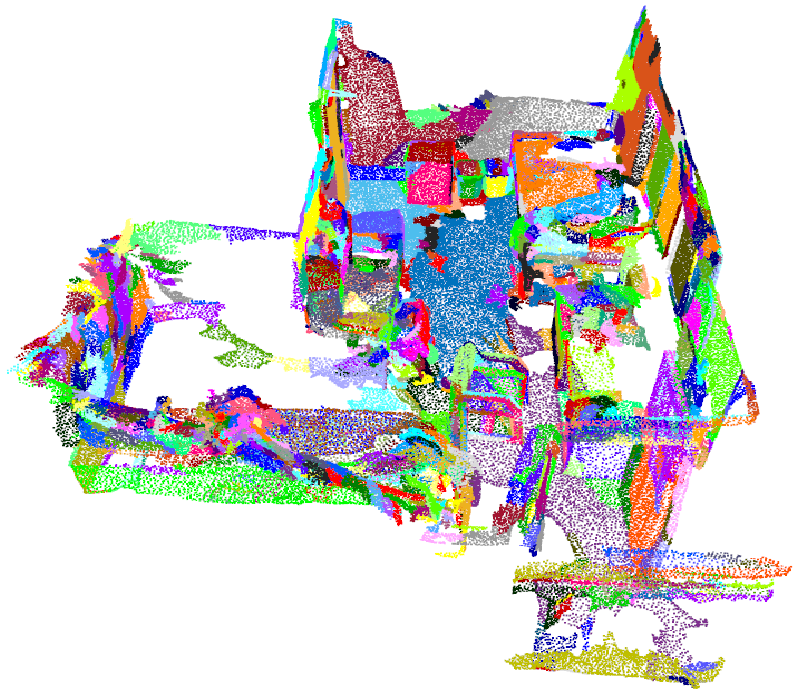} & \includegraphics[width=0.23\textwidth]{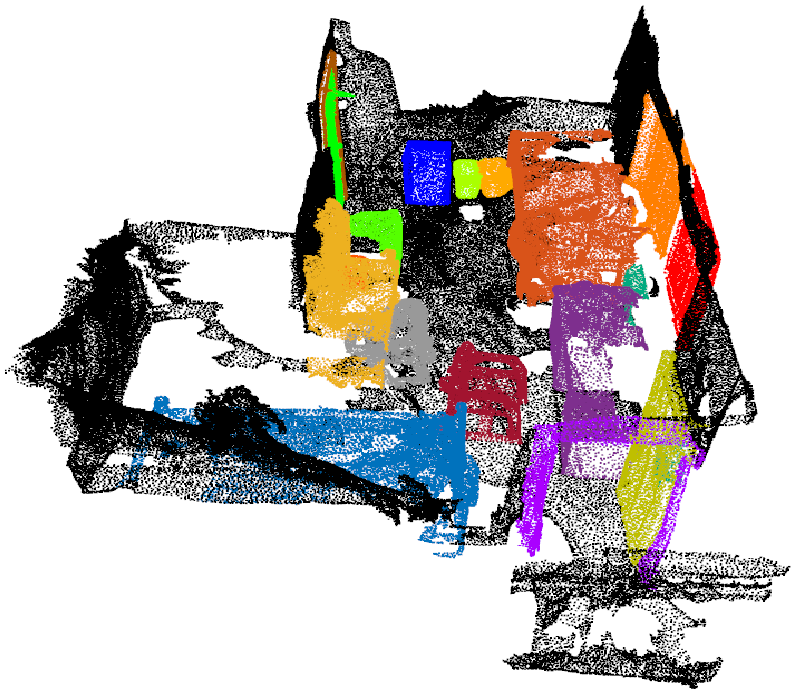} 
    \\ 
    \includegraphics[width=0.23\textwidth]{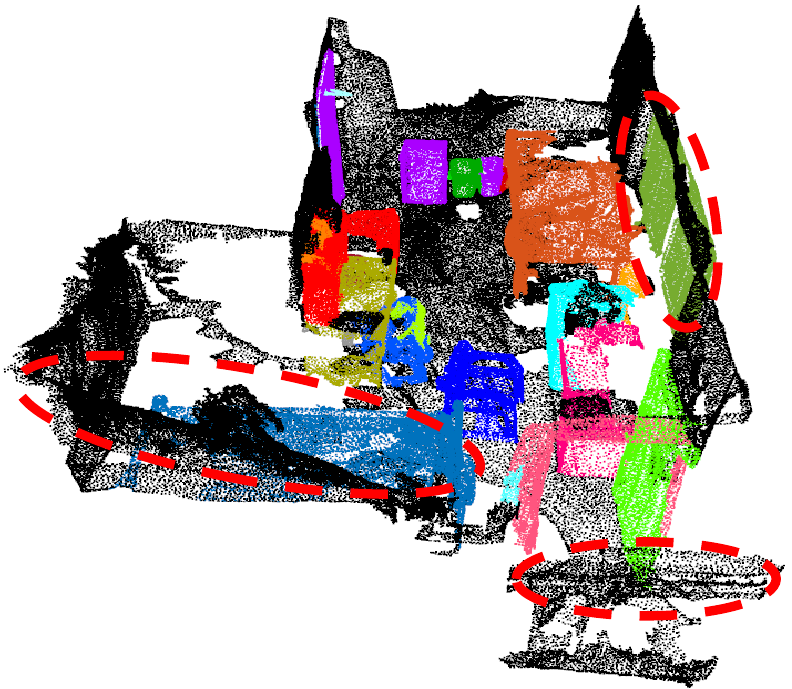} & \includegraphics[width=0.23\textwidth]{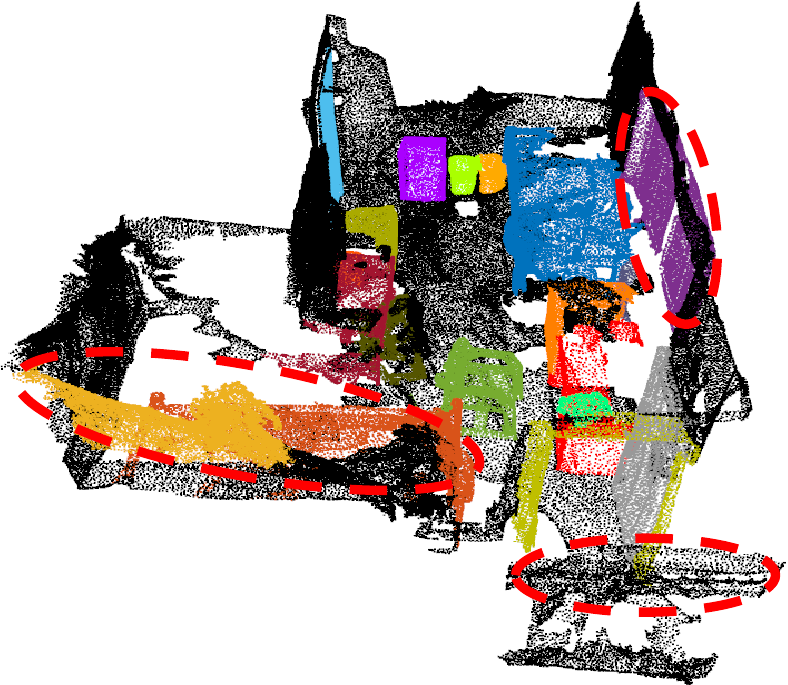} & \includegraphics[width=0.23\textwidth]{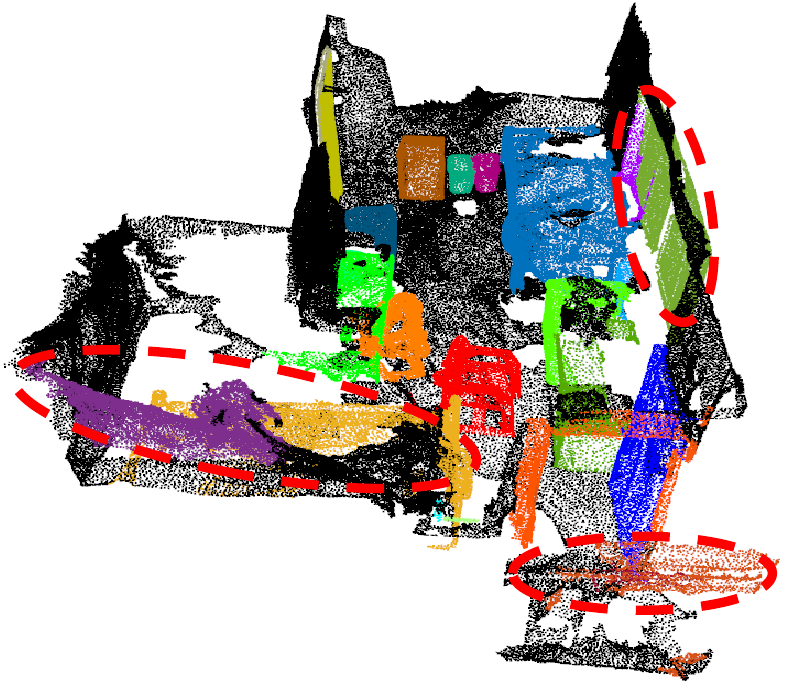} & \includegraphics[width=0.23\textwidth]{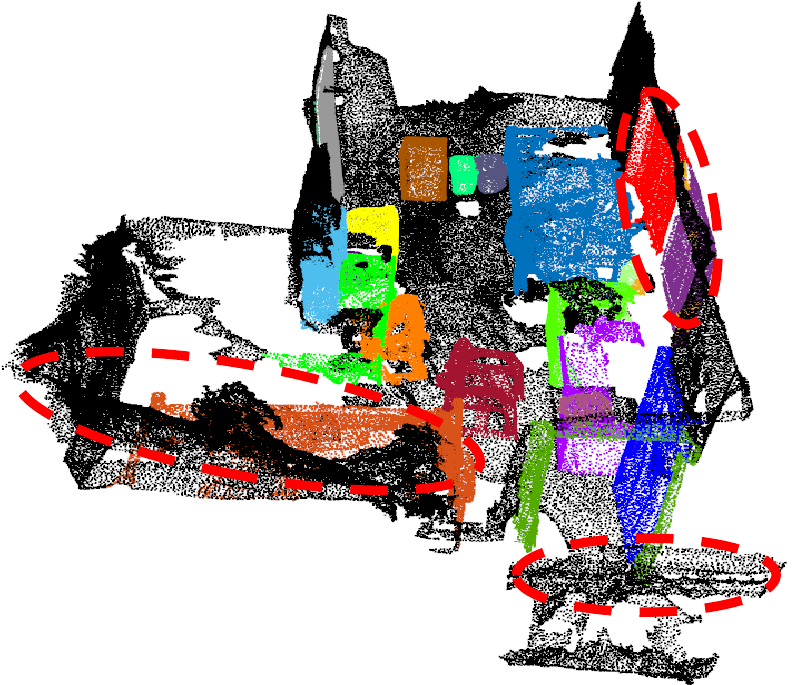}\\
    ISBNet & SPFormer & MAFT & Ours \\ \\ 
    \hline \\ \\
    Input & Low-scale Superpoint & High-scale Superpoint & Ground-truth \\
    \includegraphics[width=0.23\textwidth]{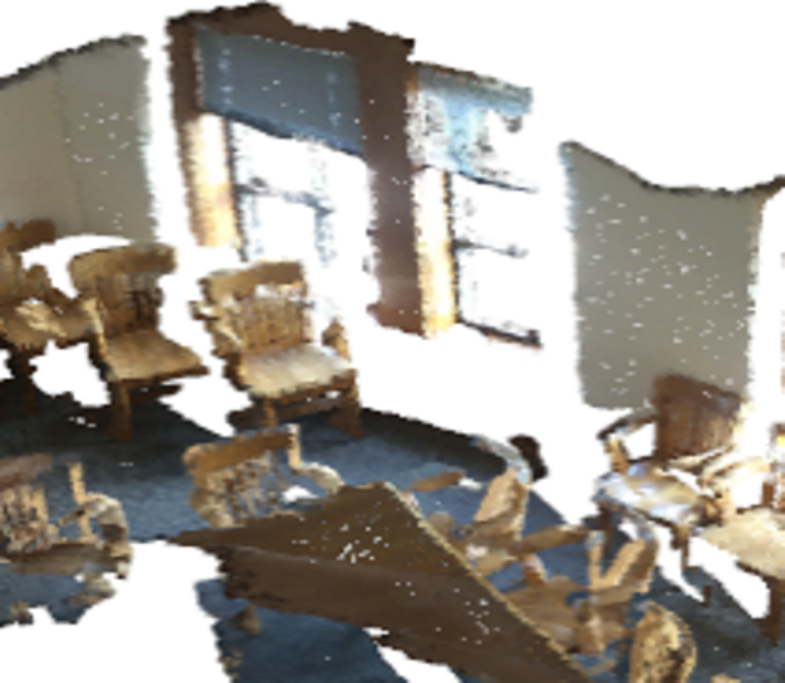} & \includegraphics[width=0.23\textwidth]{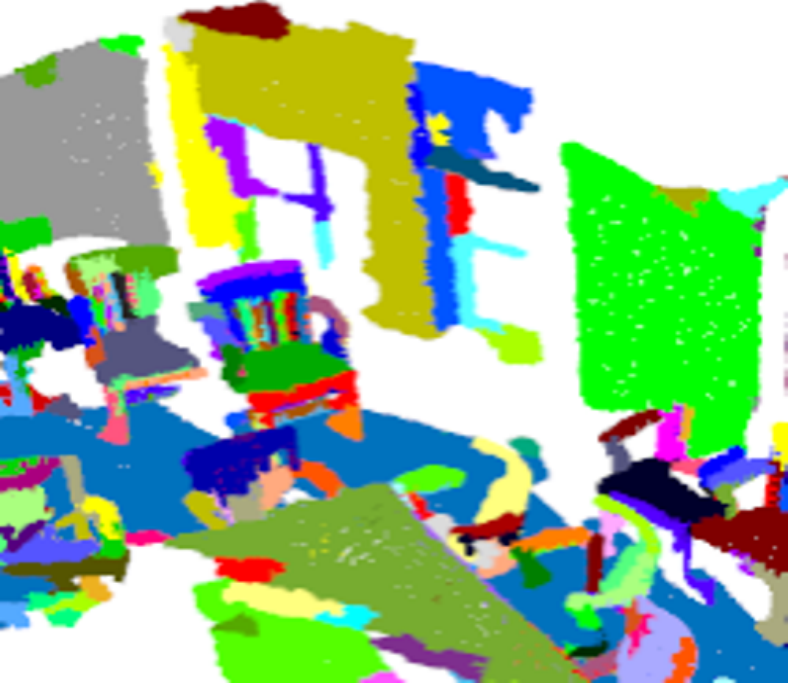} &  \includegraphics[width=0.23\textwidth]{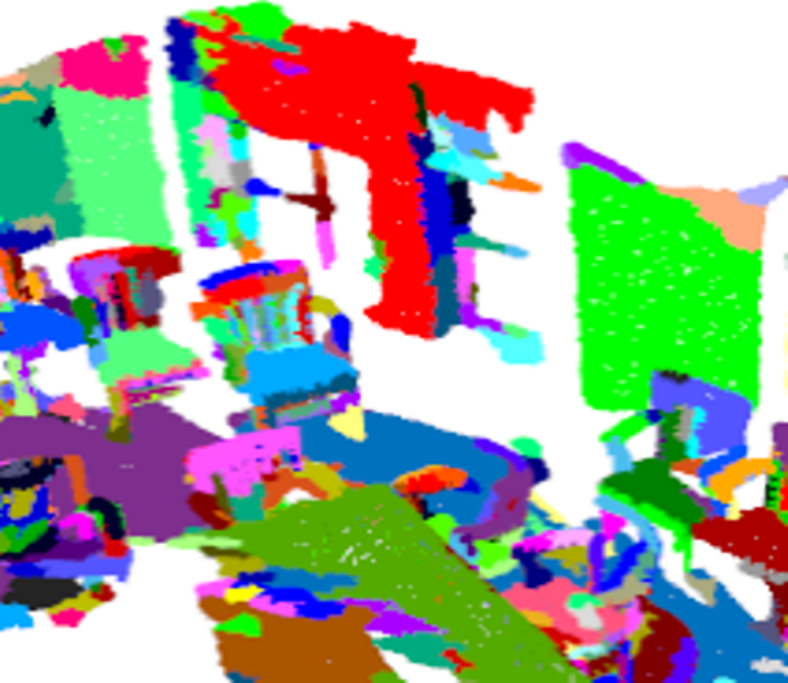} & \includegraphics[width=0.23\textwidth]{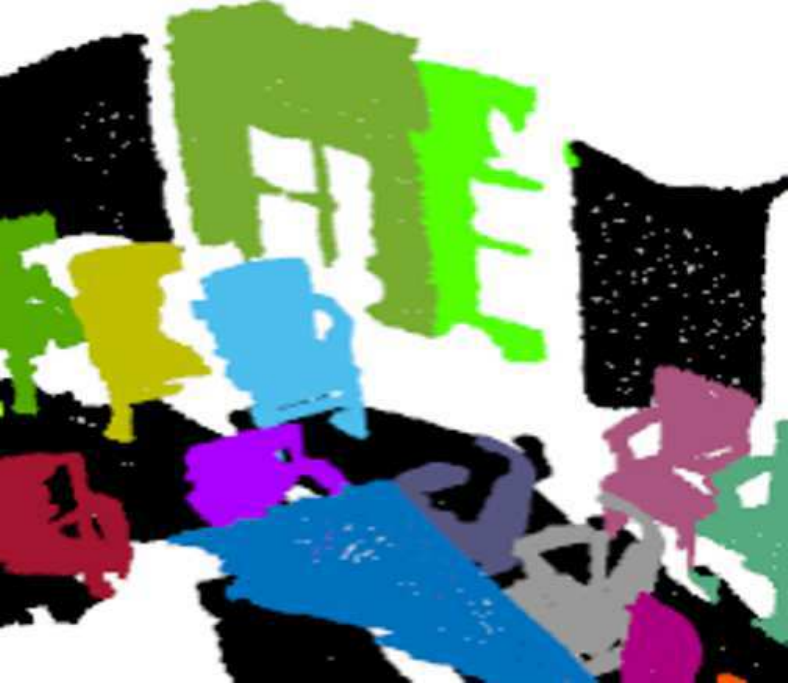} 
    \\ 
    \includegraphics[width=0.23\textwidth]{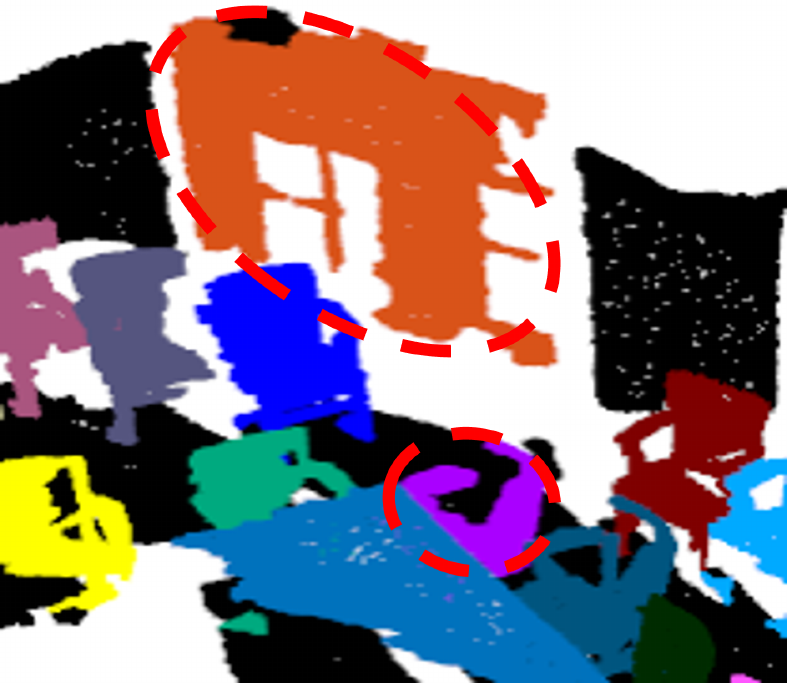} & 
    \includegraphics[width=0.23\textwidth]{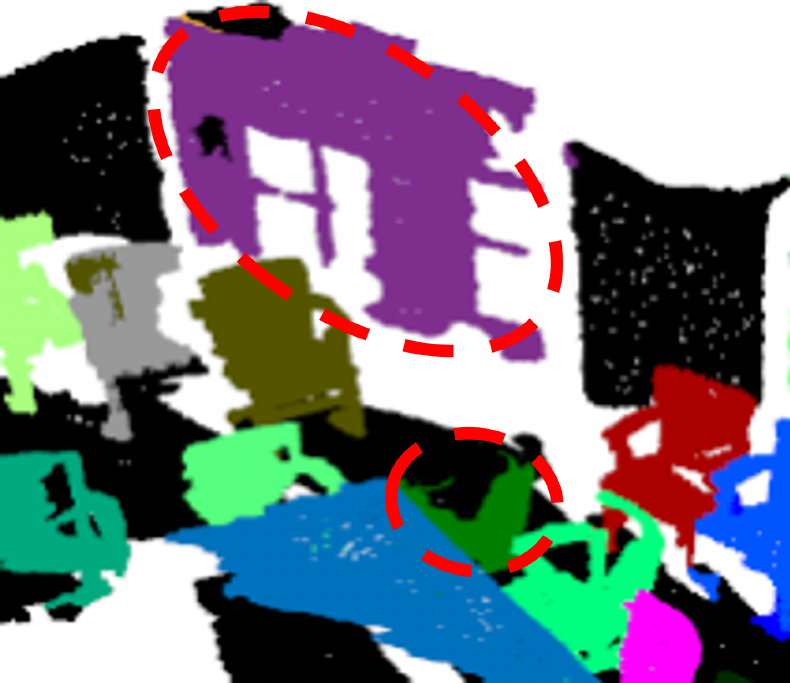} & 
    \includegraphics[width=0.23\textwidth]{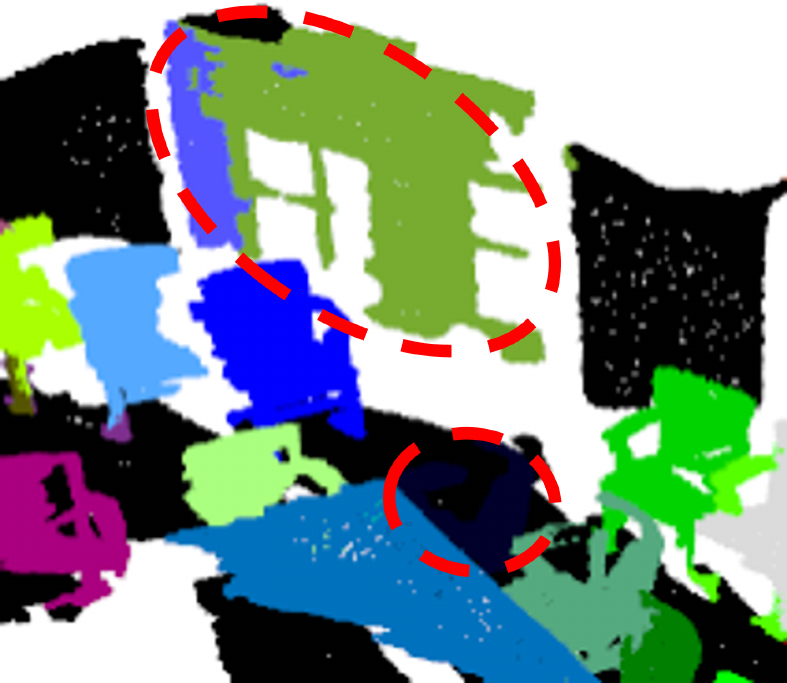} & 
    \includegraphics[width=0.23\textwidth]{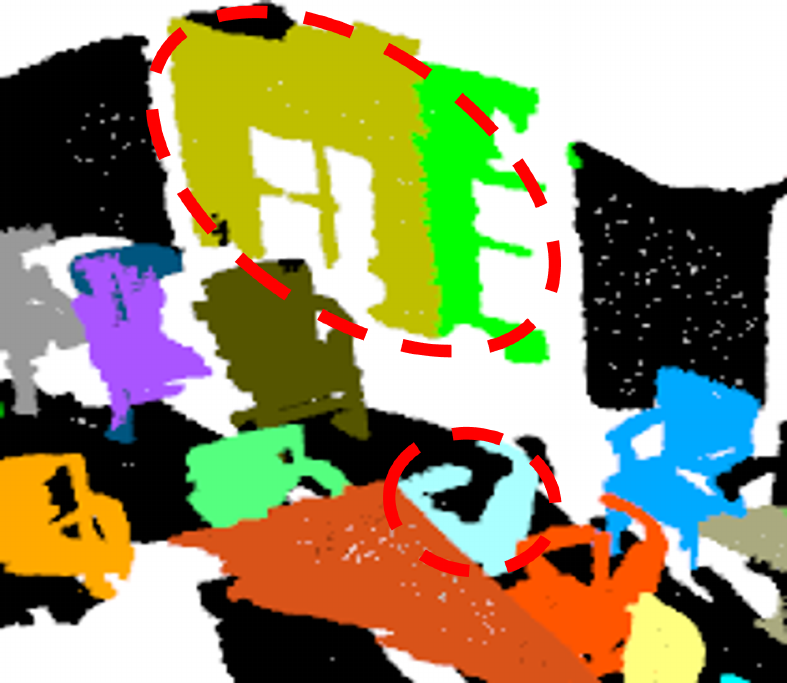} \\
    ISBNet & SPFormer & MAFT & Ours\\ 
    \end{tabular}
    \caption{Qualitative comparison of the proposed model with other methods on ScanNetV2. \vspace{-0.2cm}} \label{fig:qualitative_comparison}
\end{figure*}

\begin{figure*}[!h]
    \begin{tabular}{@{}c@{}c@{}c@{}c@{}}
    \\
    Input & Low-scale Superpoint & High-scale Superpoint & Ground-truth \\
    \includegraphics[width=0.23\textwidth]{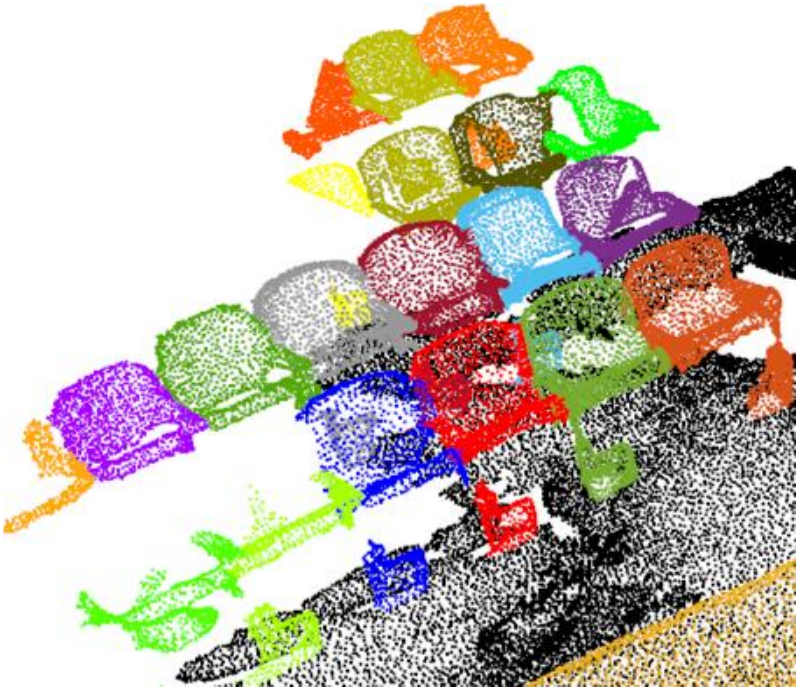} & \includegraphics[width=0.23\textwidth]{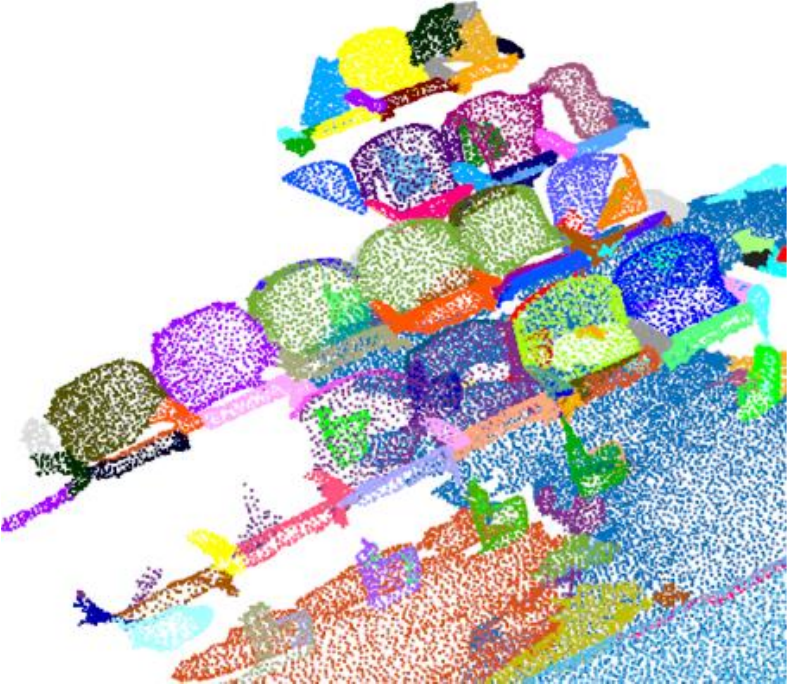} &  \includegraphics[width=0.23\textwidth]{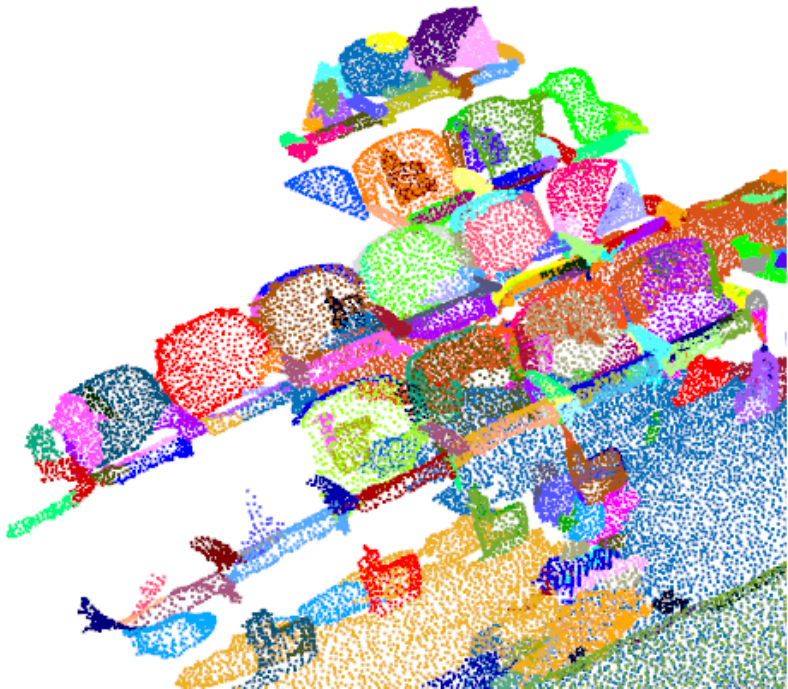} & \includegraphics[width=0.23\textwidth]{Images/Figure_6_1/Figure_6_1_ground_truth} 
    \\ 
    \includegraphics[width=0.23\textwidth]{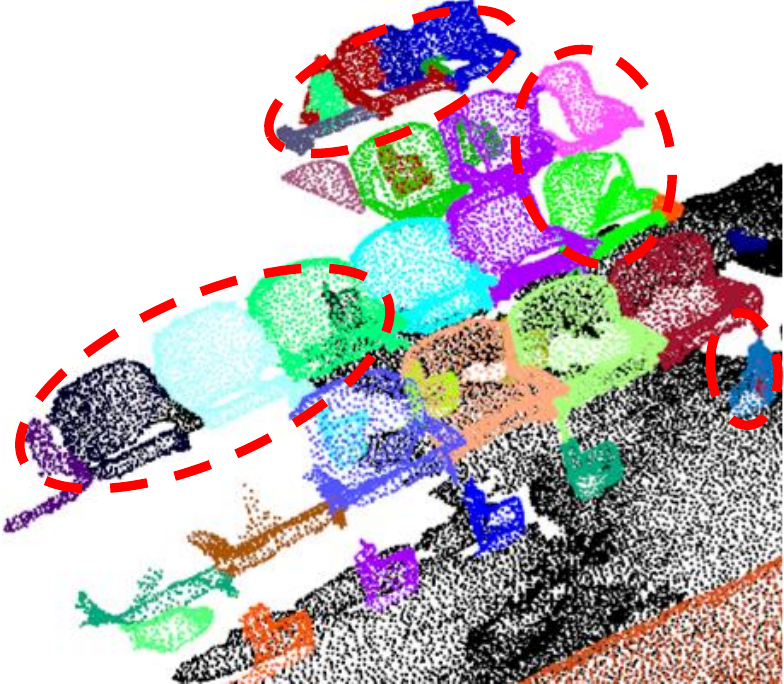} & \includegraphics[width=0.23\textwidth]{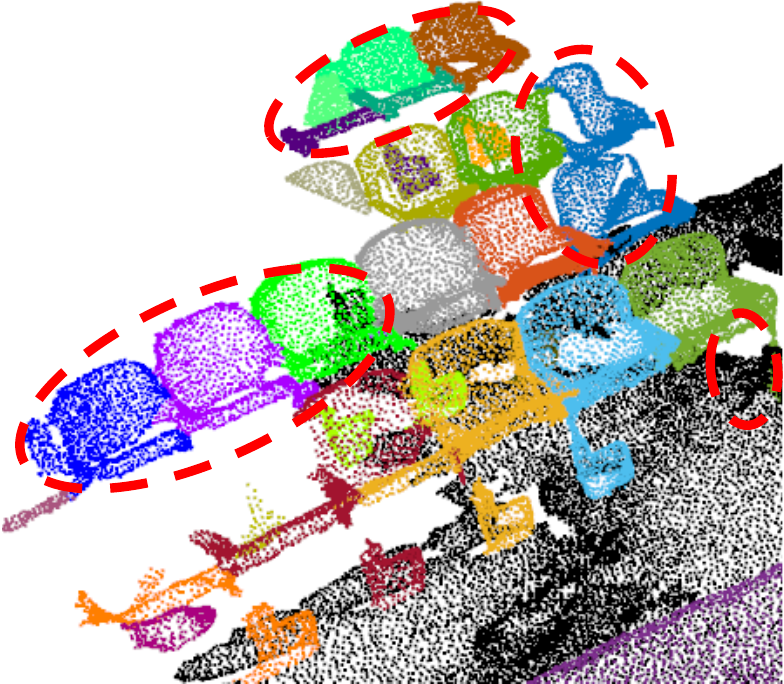} & \includegraphics[width=0.23\textwidth]{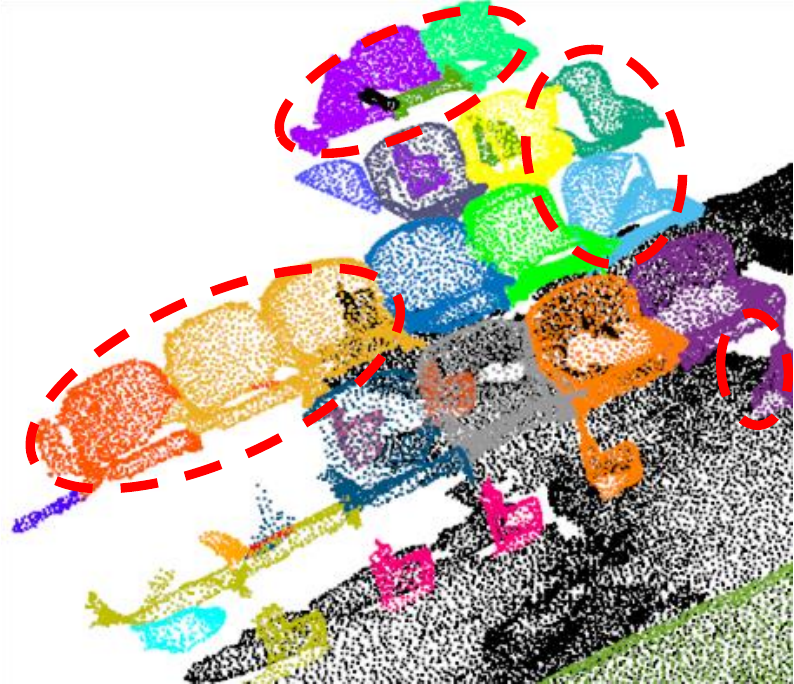} & \includegraphics[width=0.23\textwidth]{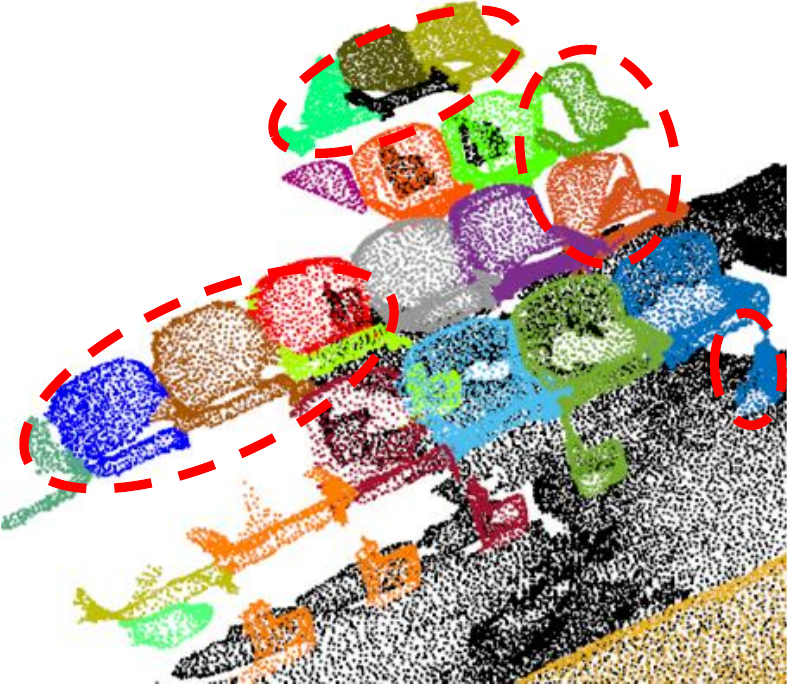}\\
    ISBNet & SPFormer & MAFT & Ours \\ \\
    \hline \\ \\
    Input & Low-scale Superpoint & High-scale Superpoint & Ground-truth \\
    \includegraphics[width=0.23\textwidth]{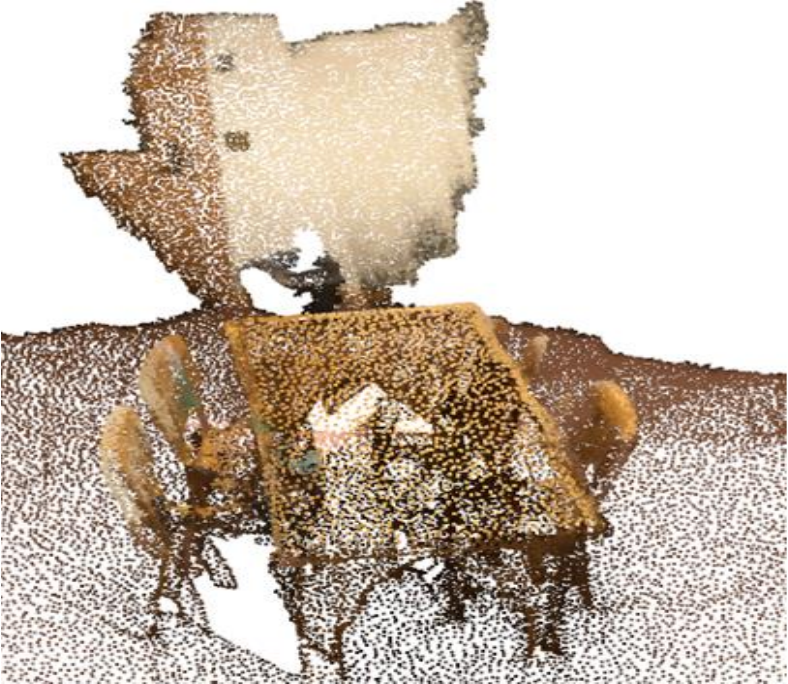} & \includegraphics[width=0.23\textwidth]{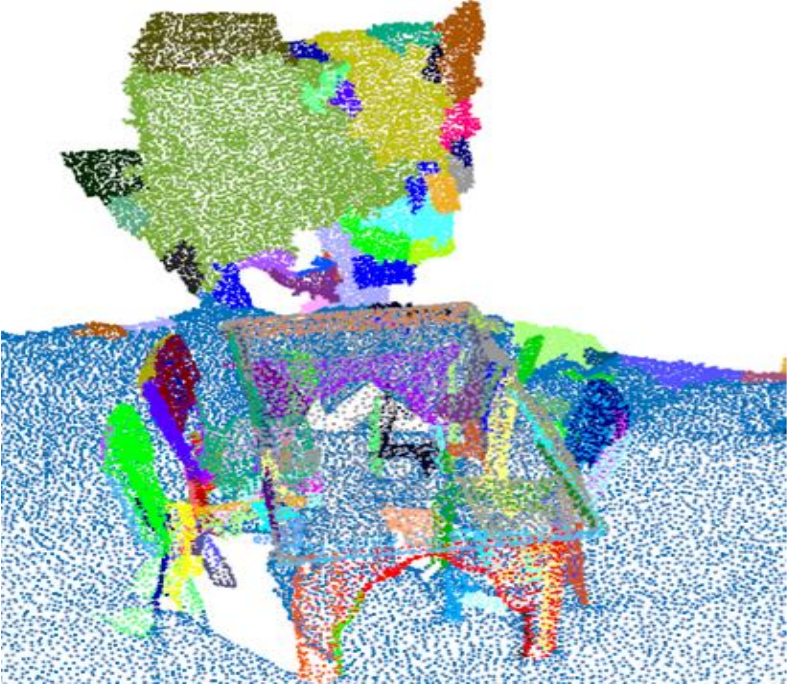} &  \includegraphics[width=0.23\textwidth]{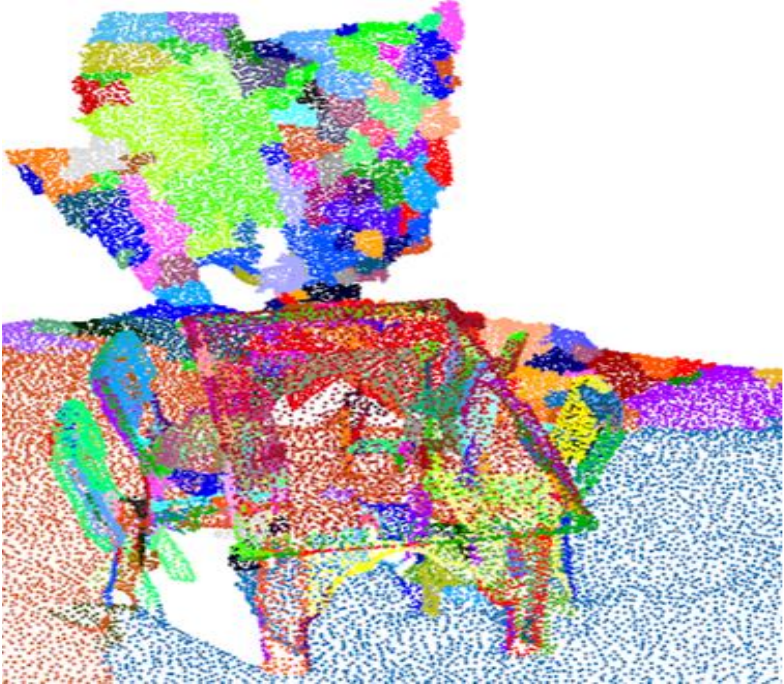} & \includegraphics[width=0.23\textwidth]{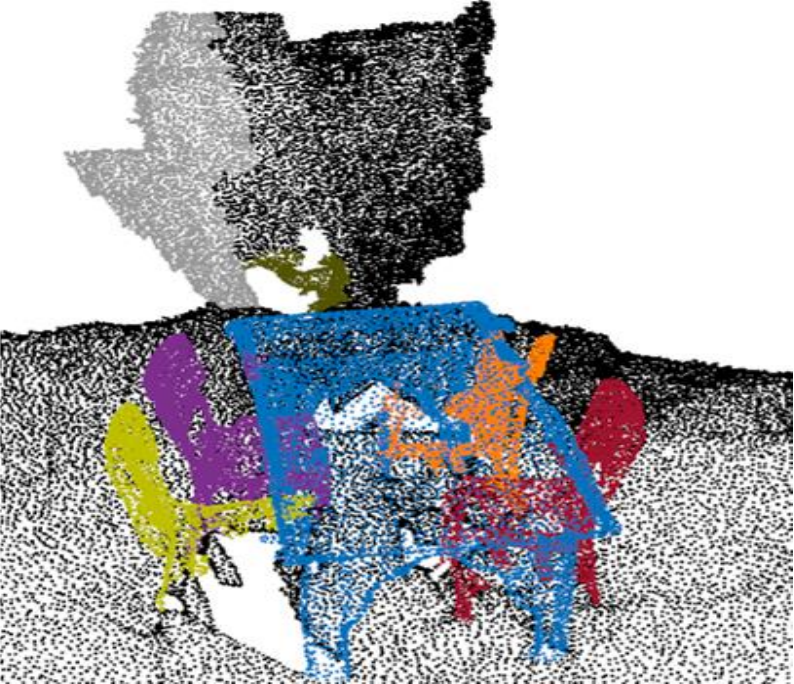} 
    \\ 
    \includegraphics[width=0.23\textwidth]{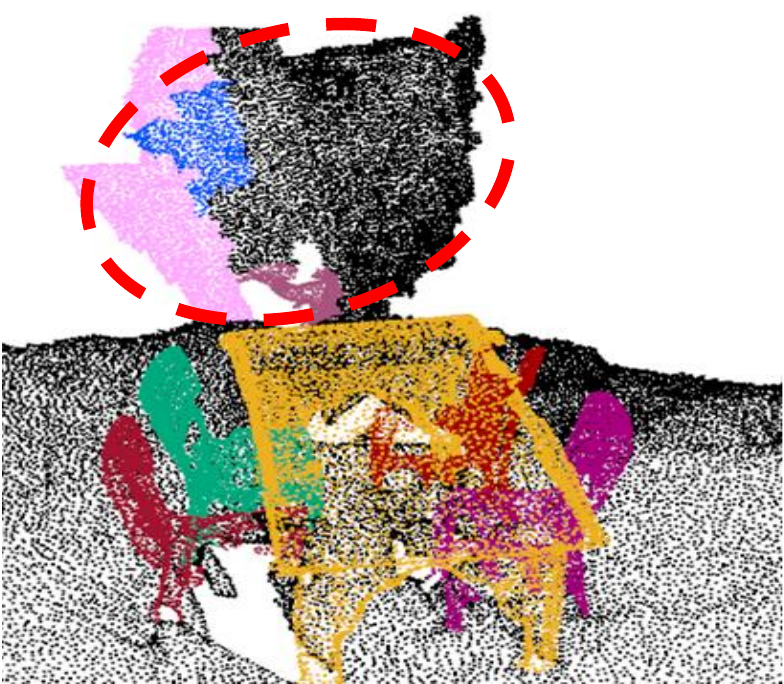} & \includegraphics[width=0.23\textwidth]{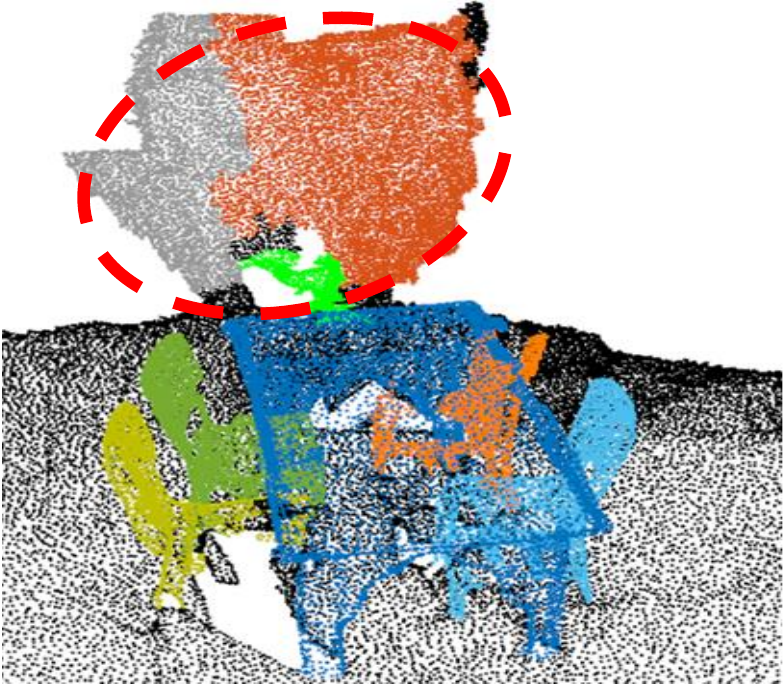} & \includegraphics[width=0.23\textwidth]{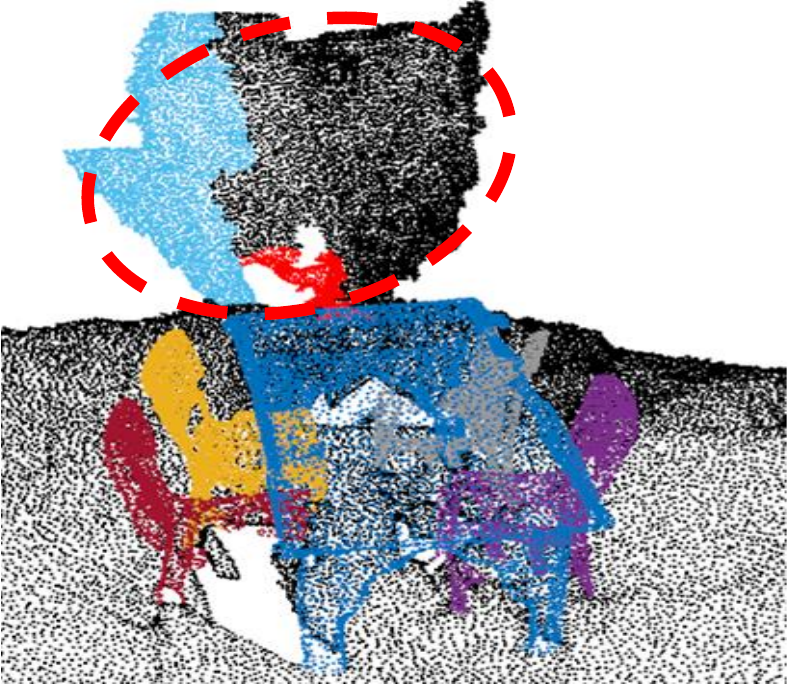} & \includegraphics[width=0.23\textwidth]{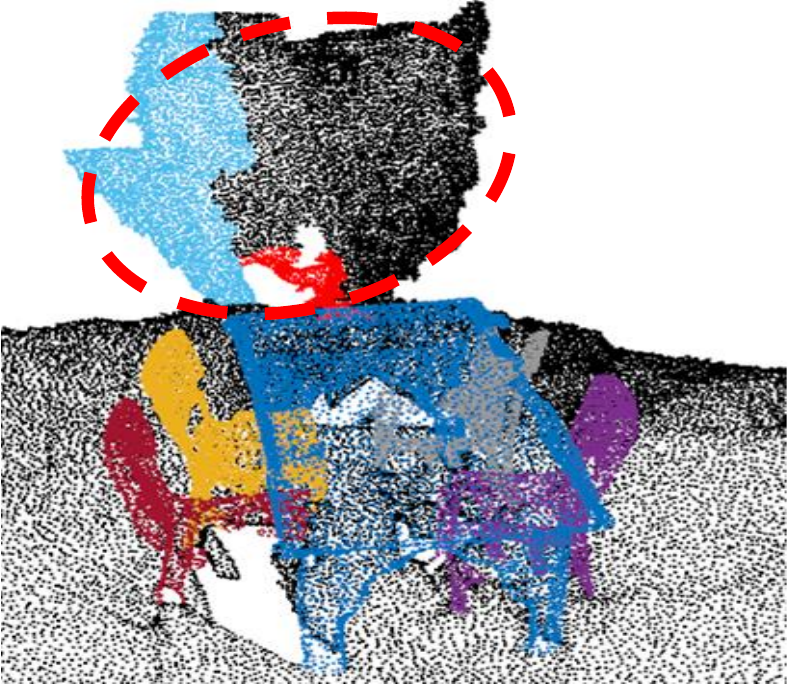}\\
    ISBNet & SPFormer & MAFT & Ours \vspace{0.2cm}\\ \\
    \end{tabular}
    \caption{Qualitative comparison of the proposed model with other methods on ScanNetV2. The proposed method mitigates inaccurate instance prediction by introducing a spatial regularizer that integrates scene-wise features and instance-specific features.} \label{fig:qualitative_comparison_6_1_2}
\end{figure*}

\textbf{ScanNetV2~\cite{dai2017scannet}.} In the top examples of \fref{fig:qualitative_comparison}, we observe that the "curtain" class for two distinct objects is often recognized as one by other methods. Additionally, these methods tend to over-segment background classes such as "wall." In the bottom examples of \fref{fig:qualitative_comparison}, the "window" class is either mispredicted or not distinguished as two separate objects due to their similarity. Our method addresses this issue by employing spatial constraints and utilizing low-scale superpoints, which aids in accurately detecting and distinguishing objects with similar features. As demonstrated in the top examples of \fref{fig:qualitative_comparison_6_1_2}, there are instances of over-segmentation observed in the chair outputs generated by the SPFormer~\cite{sun2023superpoint} and MAFT~\cite{lai2023mask} methods. Furthermore, in the bottom figure, both SPFormer~\cite{sun2023superpoint} and ISBNet~\cite{ngo2023isbnet} produce over-segmented instances with the class "wall" (background). This indicates that incorporating spatial constraints, the proposed box regularizer, improves segmentation accuracy, especially for nearby objects sharing the same label. Moreover,  the integration of low-scale superpoints with high-scale superpoints helps remove background noise, thus reducing over-segmentation in the results.

\begin{figure*}[!h]
    \begin{tabular}{@{}c@{}c@{}c@{}c@{}}
    \\
    Input & Low-scale Superpoint & High-scale Superpoint & Ground-truth \\
    \includegraphics[width=0.23\textwidth]{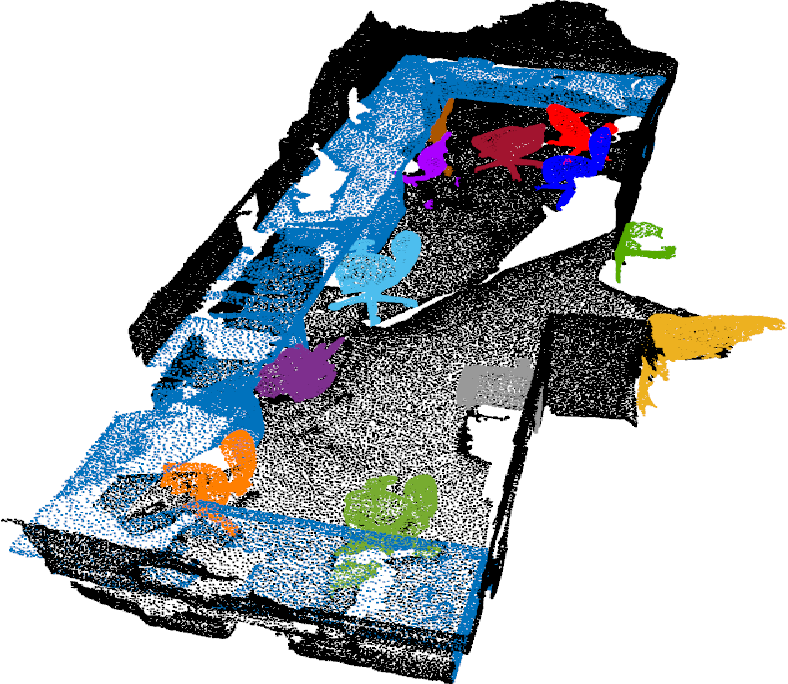} & \includegraphics[width=0.23\textwidth]{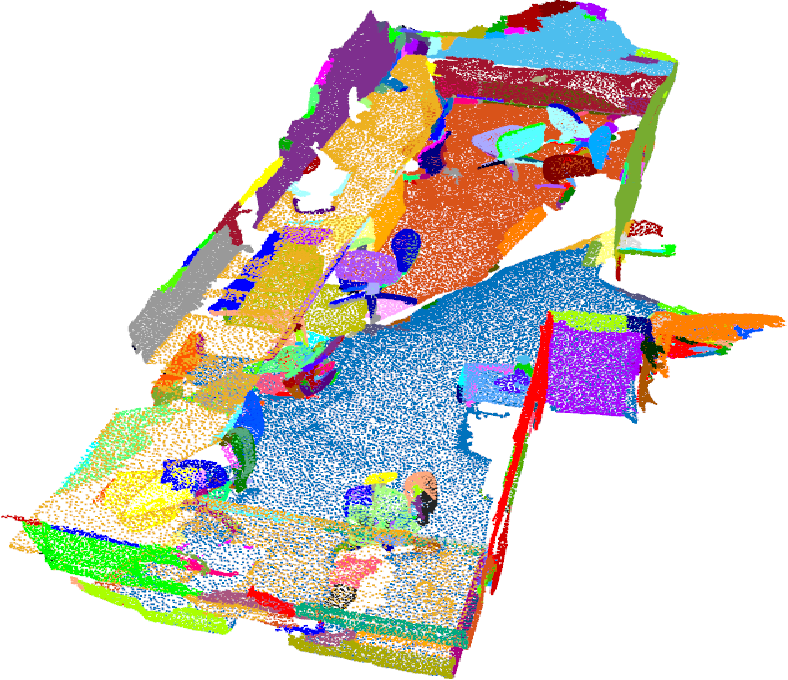} &  \includegraphics[width=0.23\textwidth]{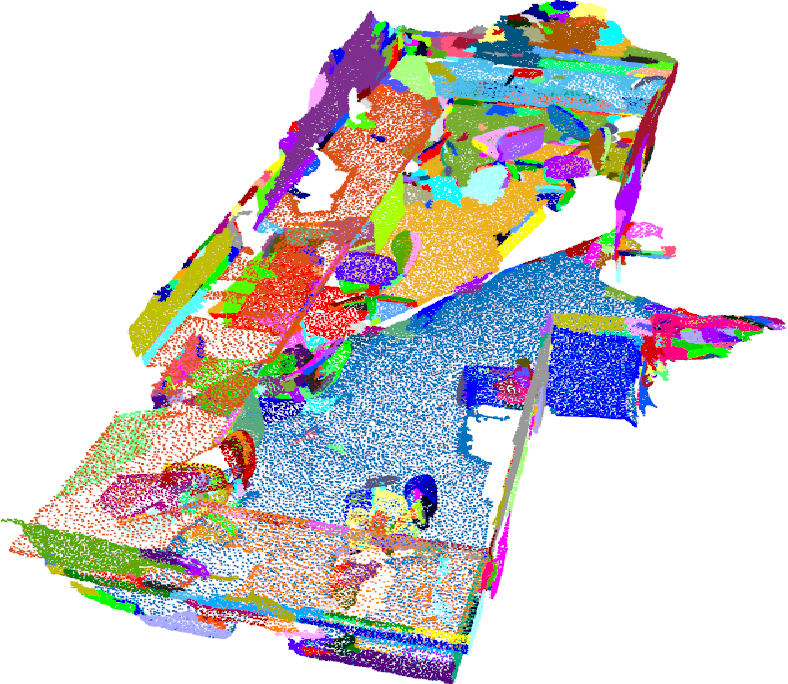} & \includegraphics[width=0.23\textwidth]{Images/Figure_6_3/Figure_6_3_ground_truth} 
    \\ 
    \includegraphics[width=0.23\textwidth]{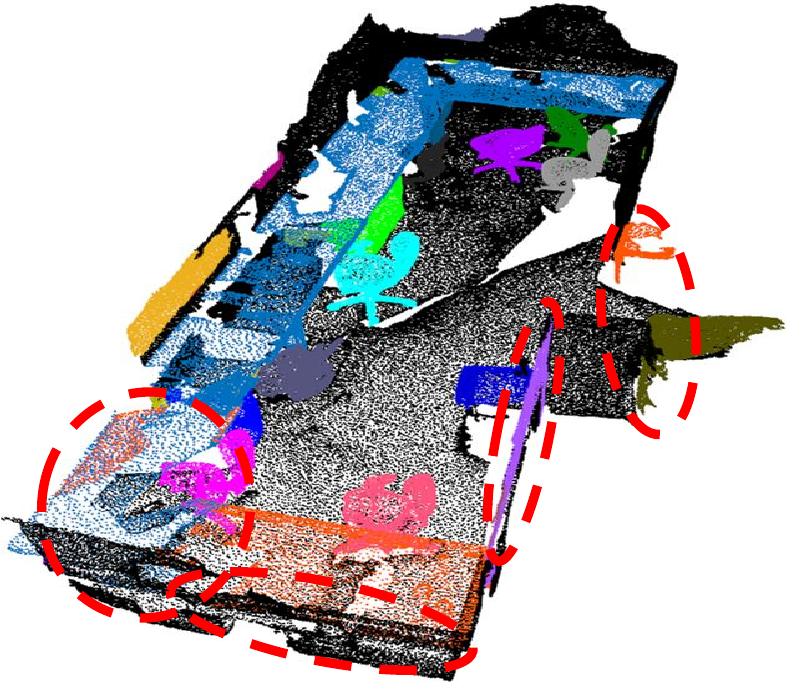} & \includegraphics[width=0.23\textwidth]{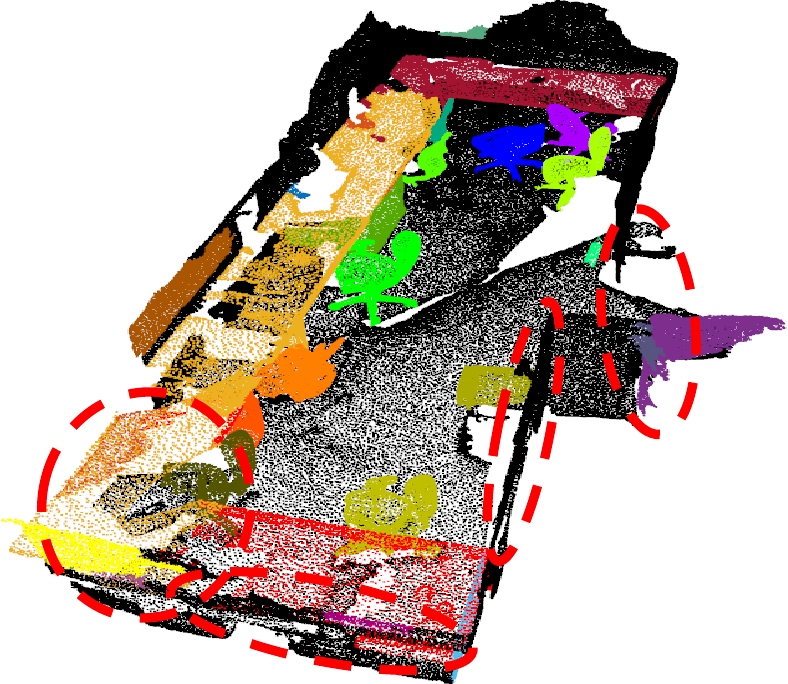} & \includegraphics[width=0.23\textwidth]{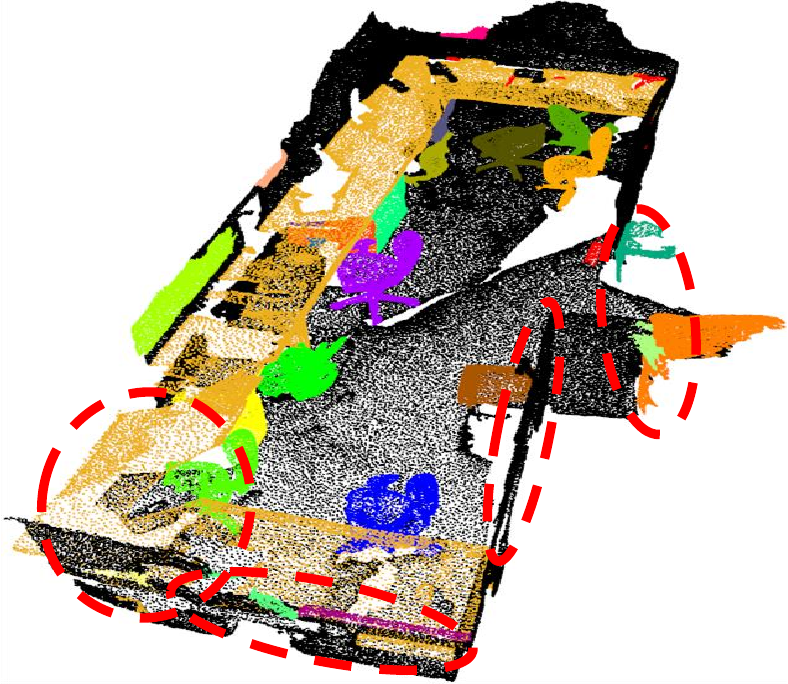} & \includegraphics[width=0.23\textwidth]{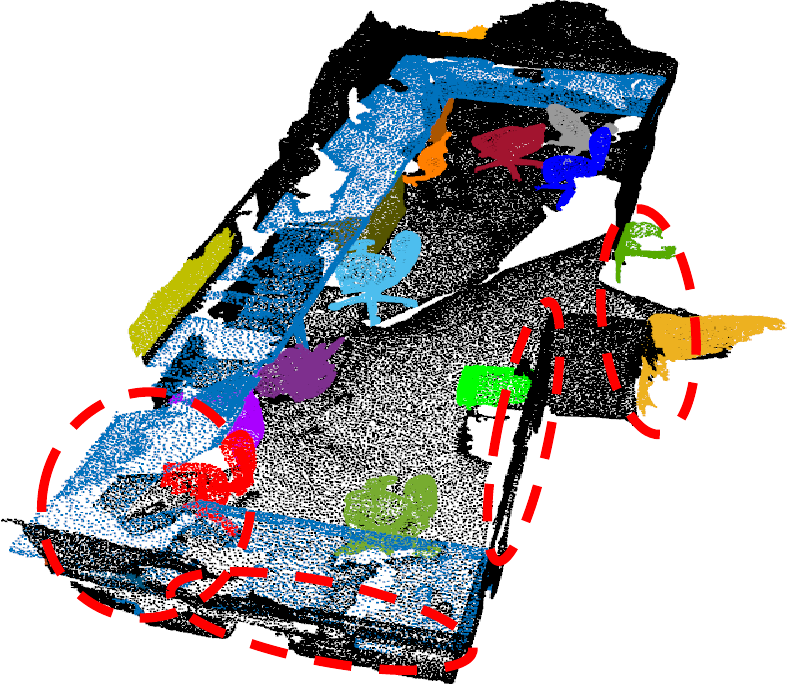}\\
    ISBNet & SPFormer & MAFT & Ours \vspace{0.2cm}\\ \\ 
    \hline \\ \\
    Input & Low-scale Superpoint & High-scale Superpoint & Ground-truth \\
    \includegraphics[width=0.23\textwidth]{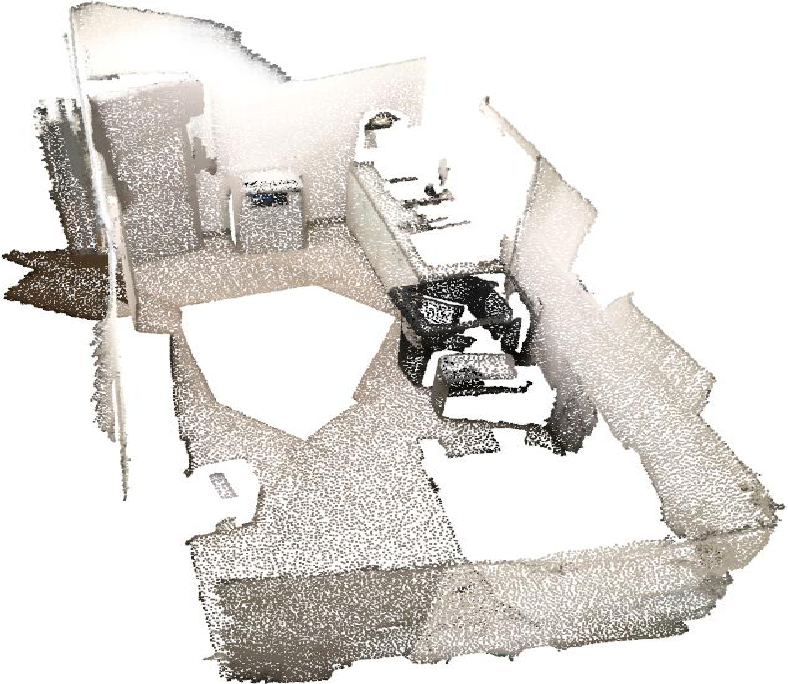} & \includegraphics[width=0.23\textwidth]{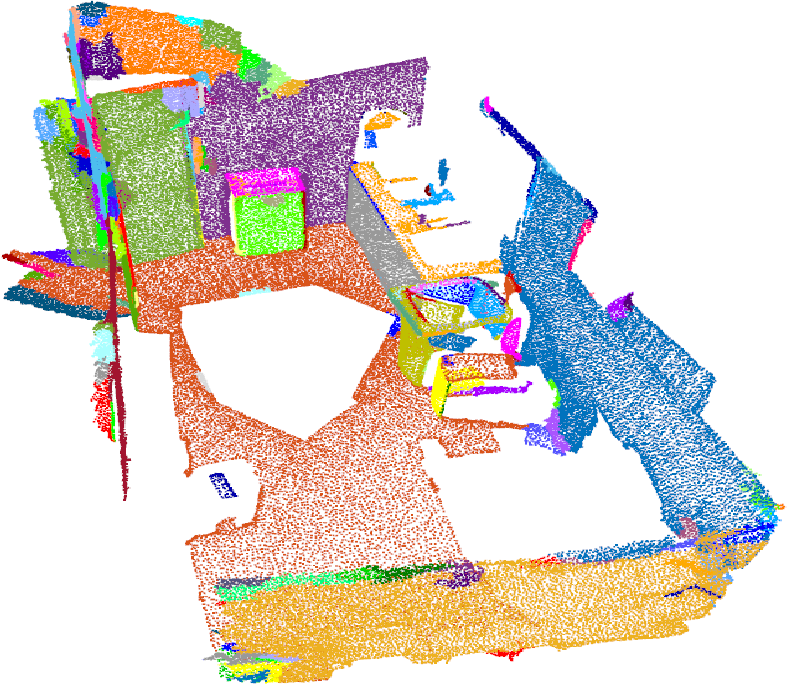} &  \includegraphics[width=0.23\textwidth]{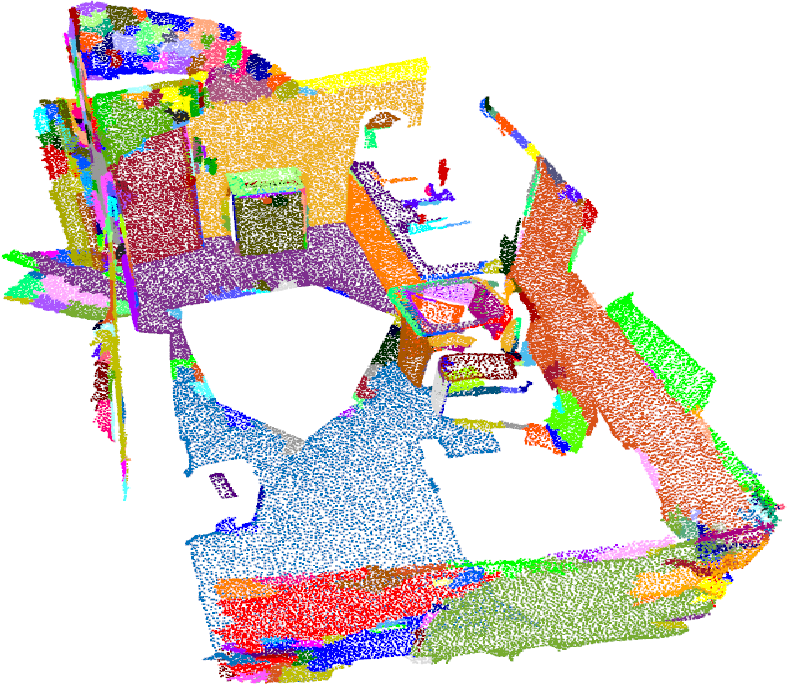} & \includegraphics[width=0.23\textwidth]{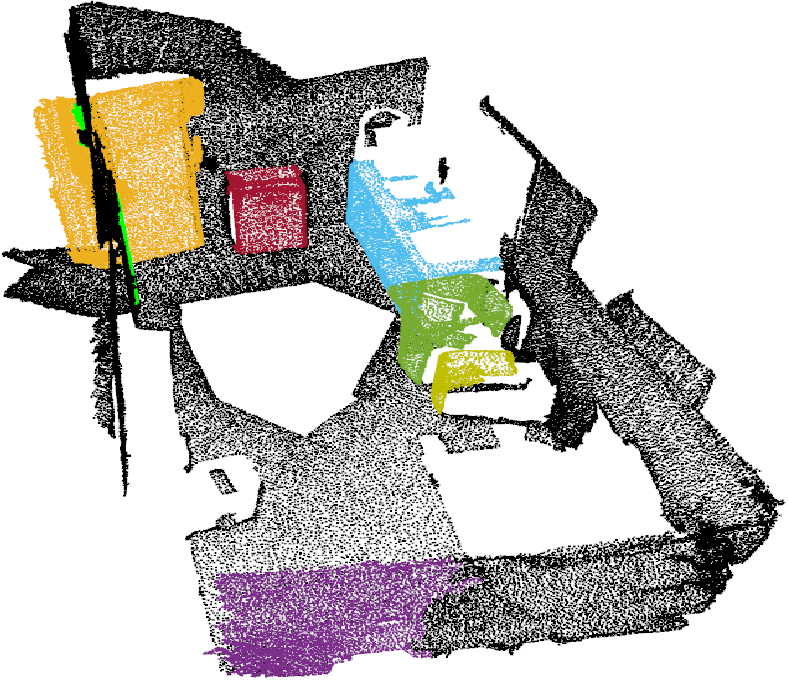} 
    \\ 
    \includegraphics[width=0.23\textwidth]{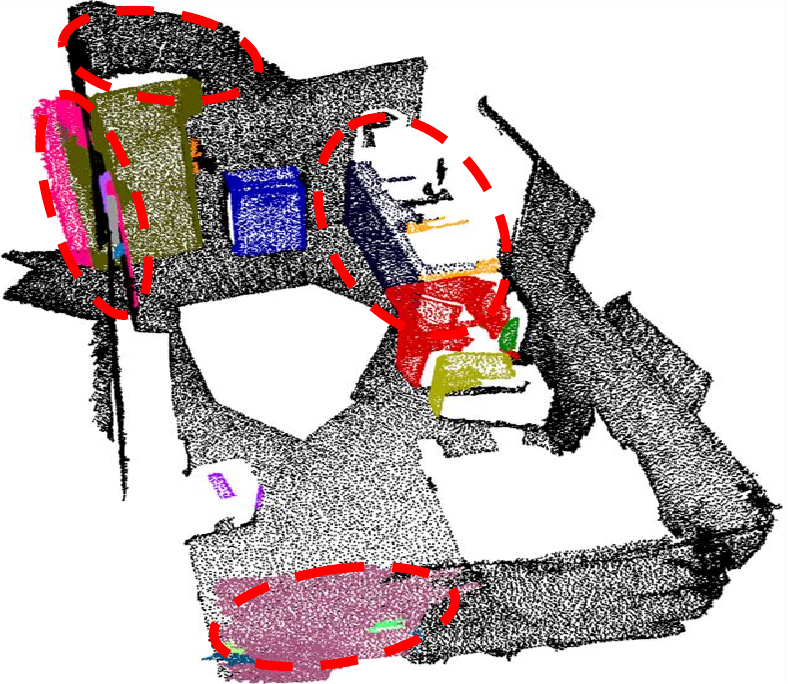} & \includegraphics[width=0.23\textwidth]{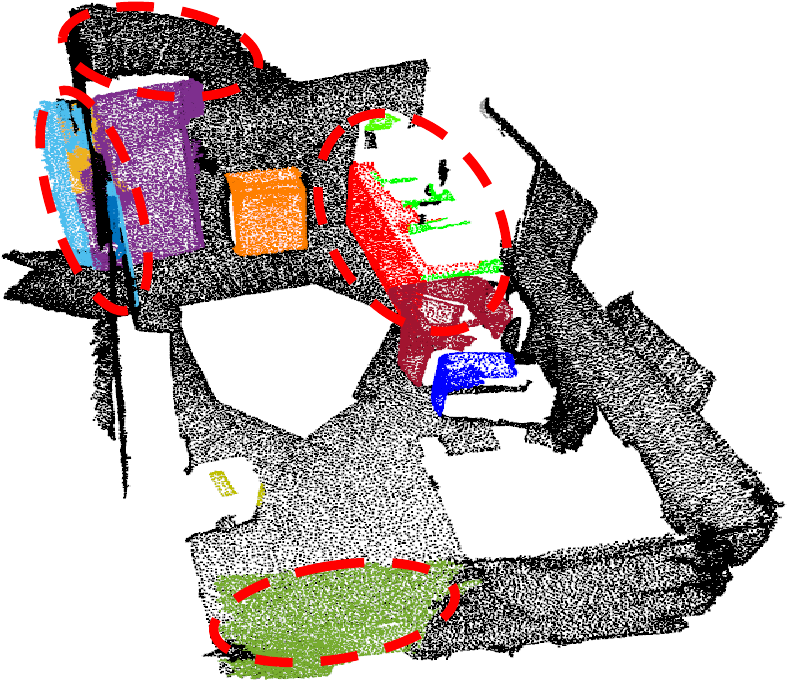} & \includegraphics[width=0.23\textwidth]{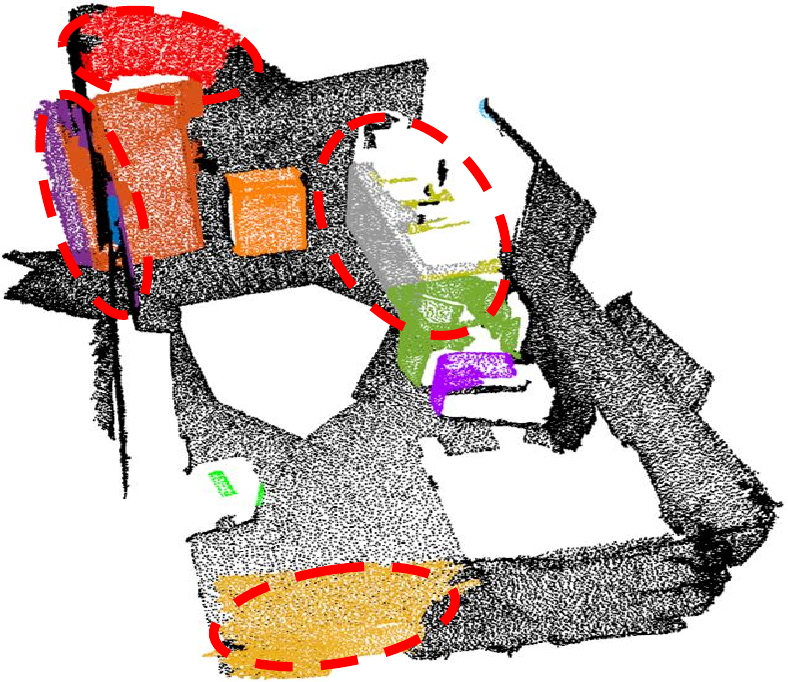} & \includegraphics[width=0.23\textwidth]{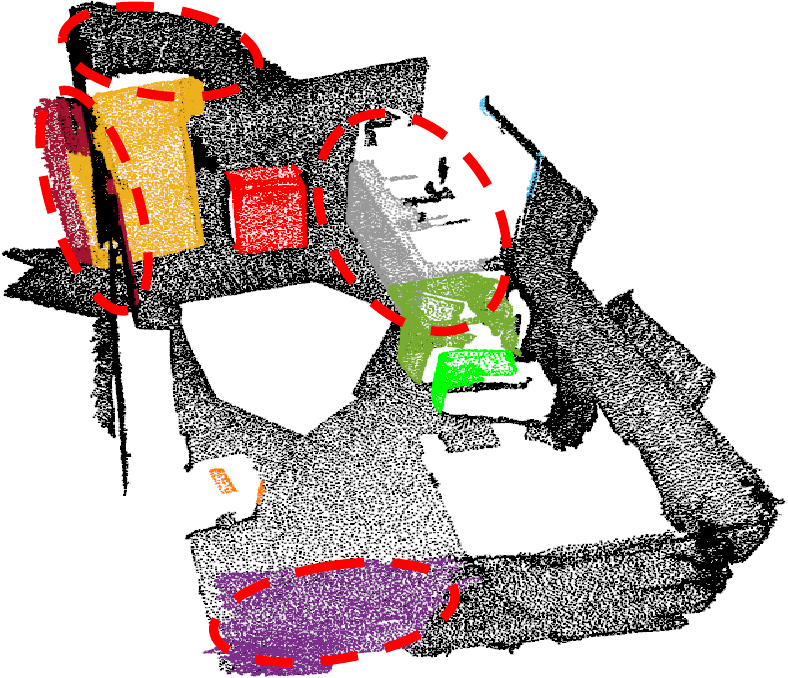}\\
    ISBNet & SPFormer & MAFT & Ours \vspace{0.2cm}\\ \\
    \end{tabular}
    \caption{Qualitative comparison of the proposed model with other methods on ScanNetV2. The proposed method overcomes over-segmentation problems by utilizing multi-scale feature representation and spatial constraints.} \label{fig:qualitative_comparison_6_3_4}
\end{figure*}

Similarly, in the top examples of Figure \ref{fig:qualitative_comparison_6_3_4}, it is observed that the class "desk" is predicted as multiple distinct objects in the outputs of ISBNet~\cite{ngo2023isbnet} and SPFormer~\cite{sun2023superpoint}. This indicates that the sampling algorithm used in ISBNet~\cite{ngo2023isbnet} is limited to large objects due to its sampling radius. In addition, relying solely on high-scale superpoints, as in the approach of SPFormer~\cite{sun2023superpoint},  proves ineffective for identifying large objects when distant superpoints lack tight connections. Over-segmentation also occurs in the "wall" (background) class for all three methods, ISBNet~\cite{ngo2023isbnet}, SPFormer~\cite{sun2023superpoint}, and MAFT~\cite{lai2023mask}.
In the bottom examples of Figure \ref{fig:qualitative_comparison_6_3_4}, all compared methods exhibit over-segmentation in the "wall" (background) and "sink" classes, whereas the proposed method alleviates this issue, demonstrating the reliability of our model's mask predictions supported by bounding box information. 

\begin{figure*}[!h]
    \begin{tabular}{@{}c@{}c@{}c@{}c@{}}
    \\
    Input & Ground-truth & ISBNet & Ours \\
    \includegraphics[width=0.23\textwidth]{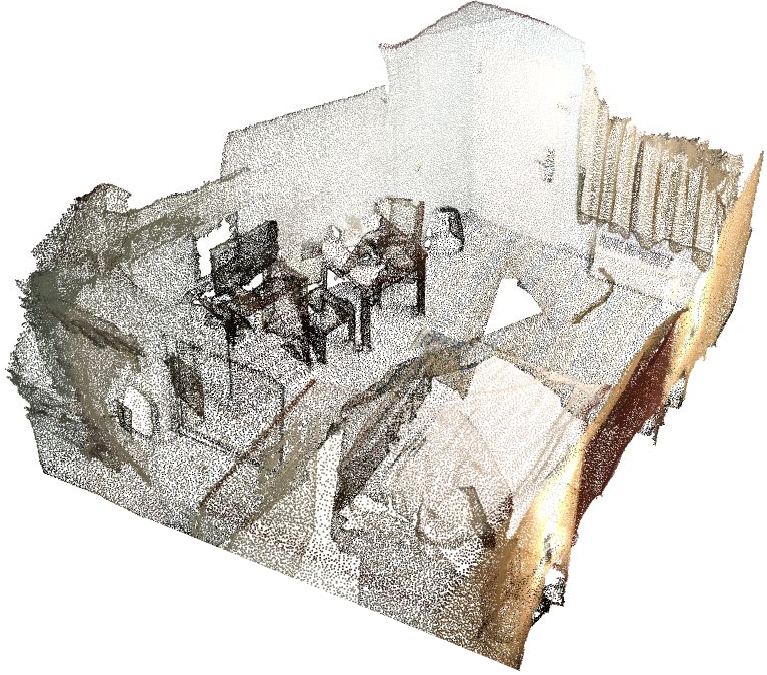} & \includegraphics[width=0.23\textwidth]{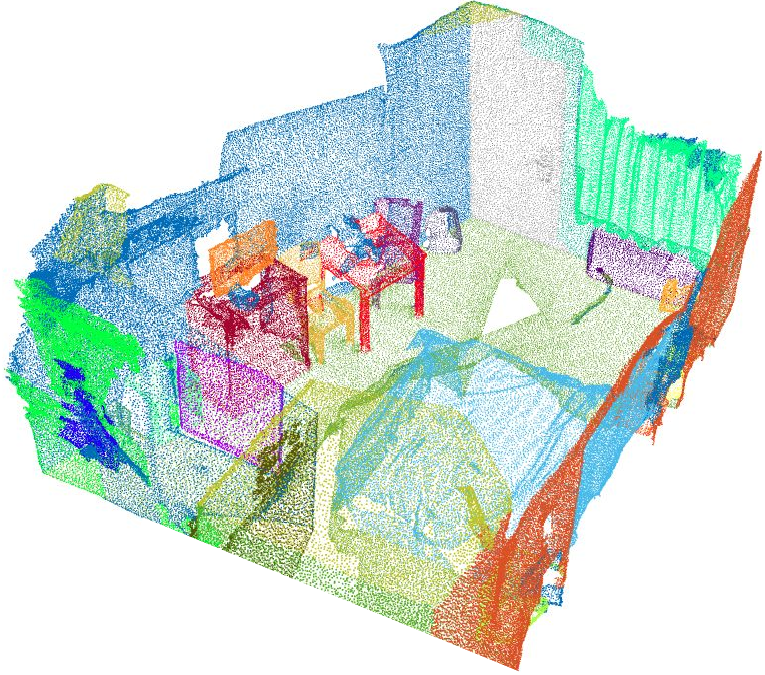} &  \includegraphics[width=0.23\textwidth]{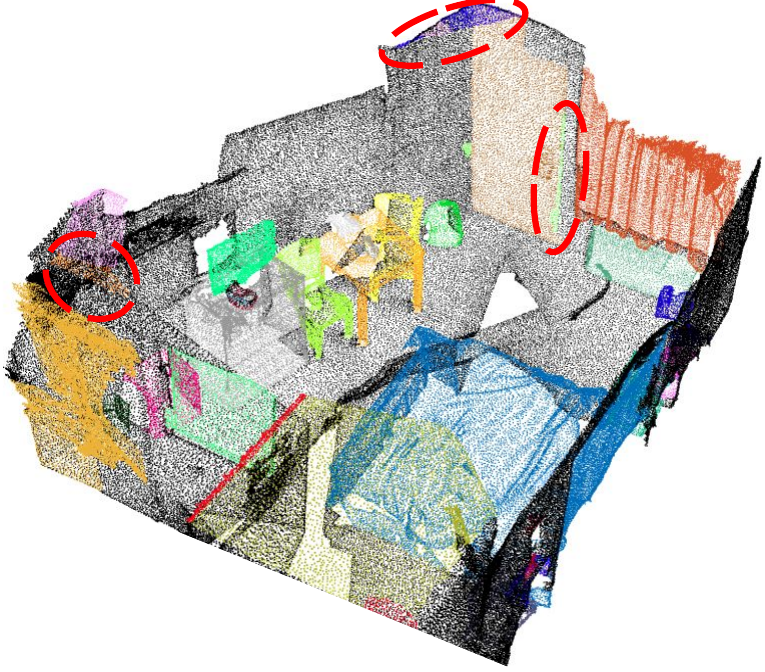} & \includegraphics[width=0.23\textwidth]{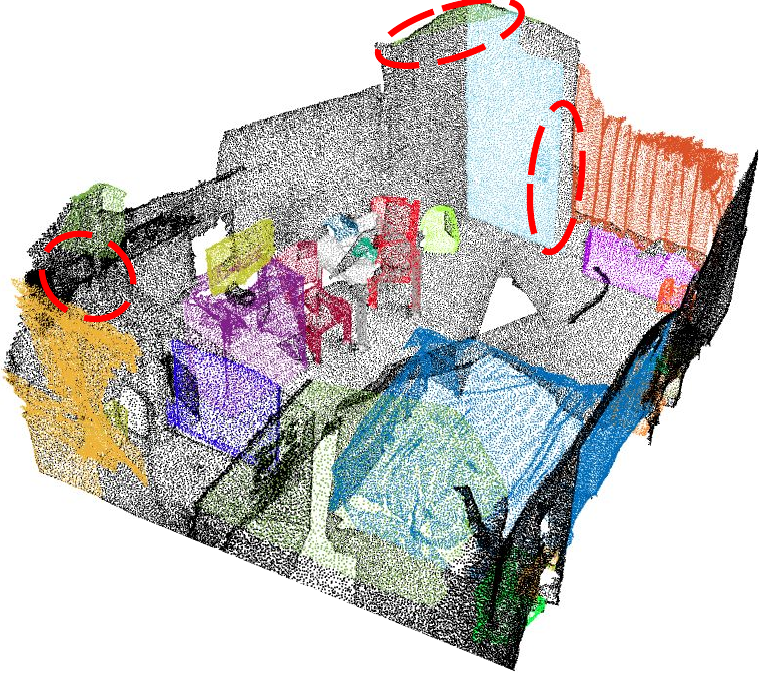} 
    \\ 
    \hline \\
    \includegraphics[width=0.23\textwidth]{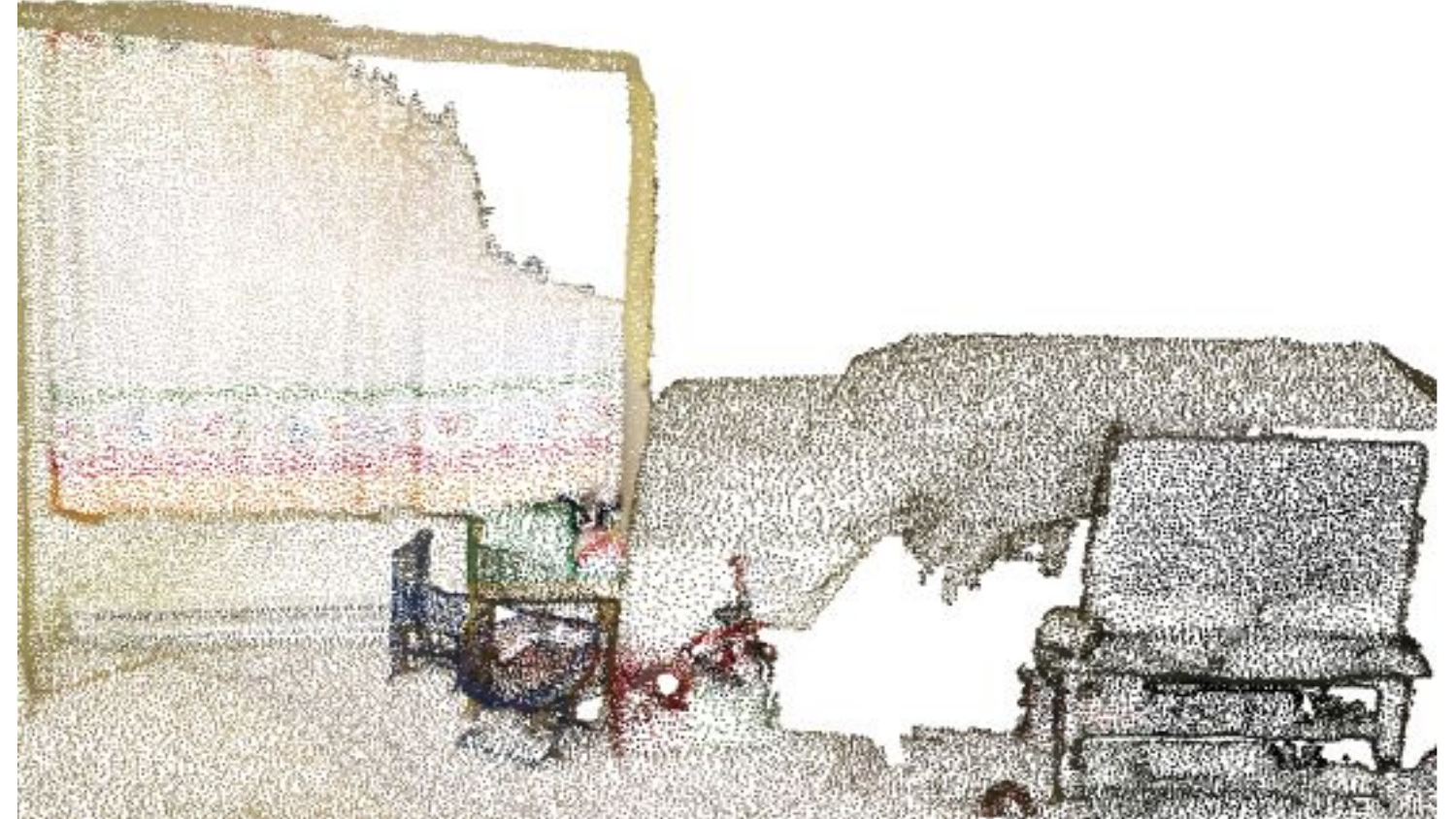} & \includegraphics[width=0.23\textwidth]{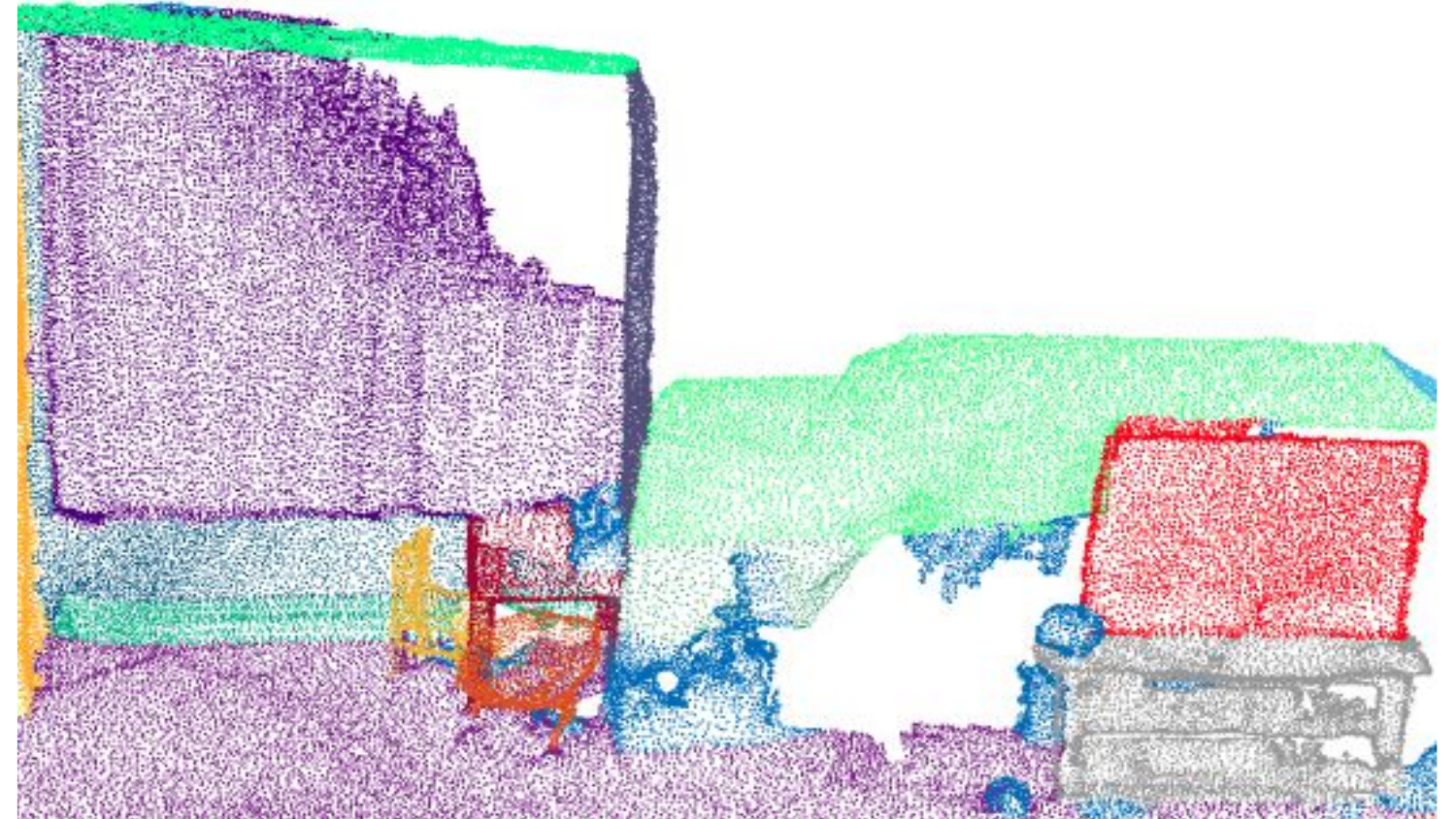} & \includegraphics[width=0.23\textwidth]{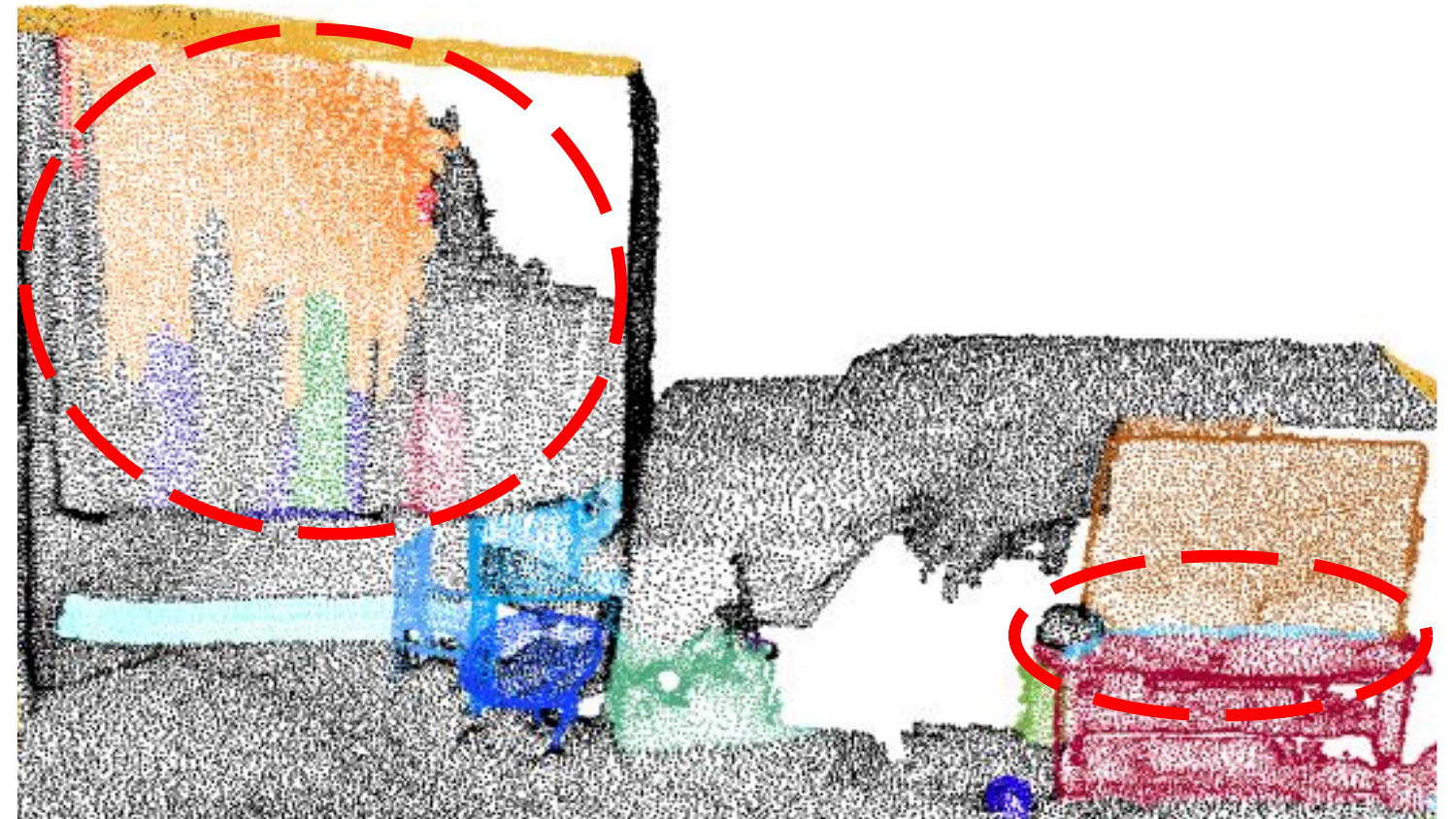} & \includegraphics[width=0.23\textwidth]{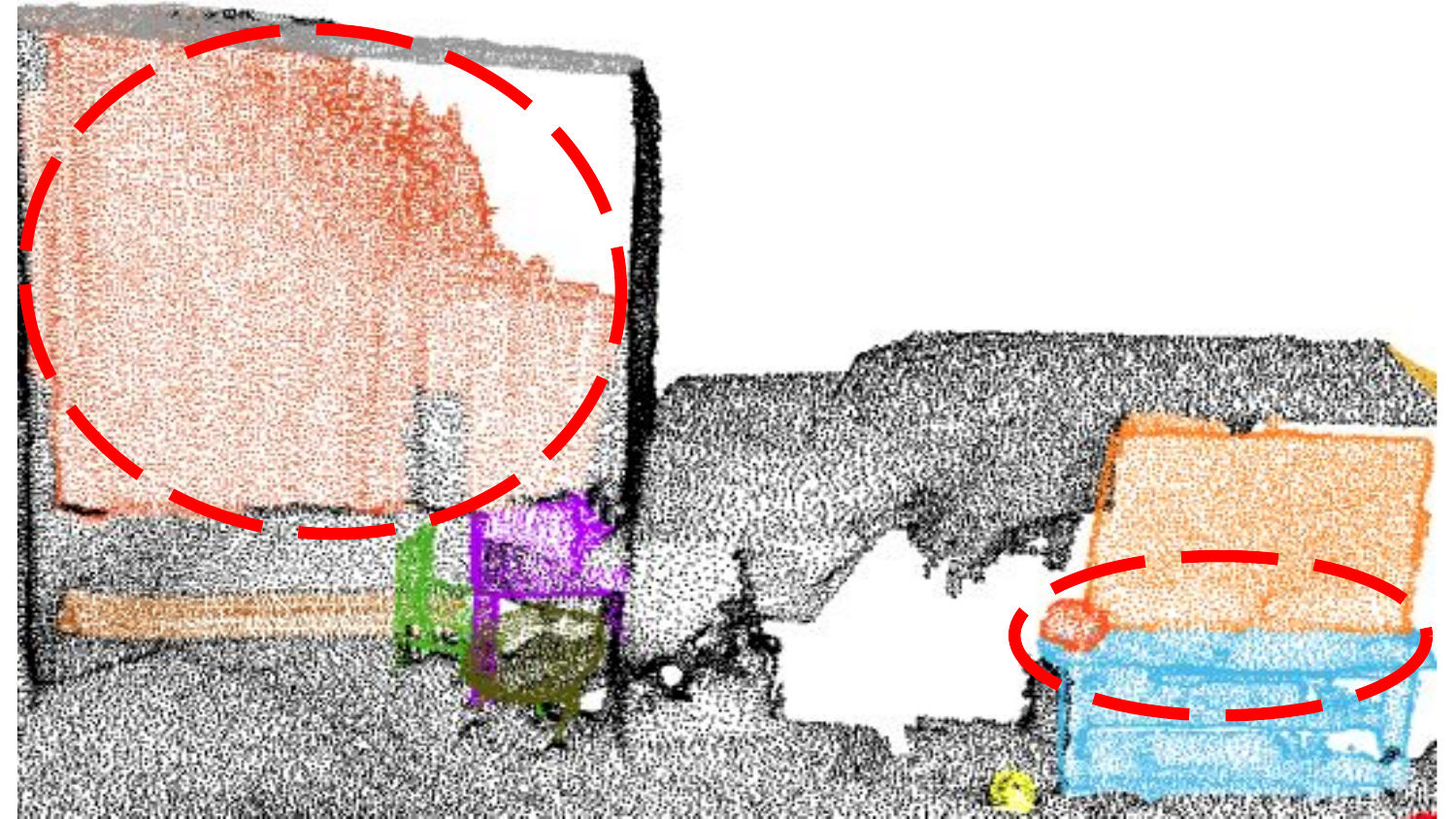}\\
    \end{tabular}
    \caption{Qualitative comparison of the proposed model with ISBNet on ScanNet200. The proposed method overcomes over-segmentation problems by utilizing multi-scale feature representation and spatial constraints.} \label{fig:qualitative_comparison_scannet200}
\end{figure*}

\textbf{ScanNet200~\cite{rozenberszki2022language} .} The over-segmentation issue is also obvious in the ScanNet200 benchmark~\cite{rozenberszki2022language}. In Figure \ref{fig:qualitative_comparison_scannet200}, we compare our results with those of ISBNet~\cite{ngo2023isbnet}. In the top examples of Figure \ref{fig:qualitative_comparison_scannet200}, ISBNet~\cite{ngo2023isbnet} show over-segmentation in the "door" class with interference from background points. Similarly, in the bottom examples of Figure \ref{fig:qualitative_comparison_scannet200}, ISBNet~\cite{ngo2023isbnet} encounters similar problems with the "curtain" and "piano" classes due to the inflexible sampling radius in its algorithm. This highlights the advantage of using multi-scale approaches, making the proposed model more adaptable in detecting objects of various sizes and thereby significantly improving performance. However, ScanNet200~\cite{rozenberszki2022language} is still a huge challenge for superpoint-based methods.


\section{Analysis of Model Complexity}
\label{app:model_complexity}
The table below summarizes the detailed architecture comparison of our method with MAFT~\cite{lai2023mask}, which has the smallest parameters among the existing methods. 
We created a lighter decoder by incorporating a loss function into the global and local feature fusion strategy of the proposed box regularizer, avoiding the use of positional embeddings in MAFT~\cite{lai2023mask}, which would have increased the model's complexity. \vspace{-0.3cm}

\begin{table}[!h]
    \caption{Comparison of the number of decoder parameters and instance prediction \vspace{-0.3cm}}
    \label{tab:model_complexity_table}
    \renewcommand{\tabcolsep}{2mm}{ {\renewcommand{\arraystretch}{0.15}
    \begin{tabular}{@{}l@{~}lccc@{}}
        \toprule
         Component & Sub-component & MAFT & Ours \\ \midrule
         Decoder & Input Projection & 8,960 & 8,960 & \\ 
         & Query-related & 663,552 & 76,800 & \\
         & Cross-attention & 2,434,304 & 1,582,080 & \\
         & Self-attention & 2,368,512 & 1,582,512 & \\
         & Feedforward & 3,162,624 & 3,156,480 & \\
         & Others & 265,651 & 512 & \\ 
         Prediction Head & & 277,527 & 416,889 & \\ \midrule
         Total parameters & & 9,181,130 & 6,824,233  \\ \bottomrule 
    \end{tabular} \vspace{-0.2cm}}}
\end{table} 

\end{document}